\pdfoutput=1
\documentclass[journal]{IEEEtran}
\usepackage{times}
\usepackage{epsfig}
\usepackage{graphicx}
\usepackage{amsmath}
\usepackage{amssymb}
\usepackage{cite}
\usepackage{multirow}
\usepackage{color}
\usepackage{algorithmic}
\usepackage{array}
\usepackage{stfloats}
\usepackage{url}
\usepackage{times}
\usepackage{algorithm}
\usepackage{subfig}
\usepackage{multirow}
\usepackage{color,xcolor}
\usepackage[normalem]{ulem}
\useunder{\uline}{\ul}{}

\begin{document}
%

\title{CircleNet: Reciprocating Feature Adaptation for Robust Pedestrian Detection}

%

\author{Tianliang~Zhang,~\IEEEmembership{Student Member,~IEEE,} Zhenjun Han,~\IEEEmembership{Member,~IEEE,} Huijuan Xu,~\IEEEmembership{Member,~IEEE,} 
\\Baochang Zhang,~\IEEEmembership{Member,~IEEE,} Qixiang~Ye,~\IEEEmembership{Senior~Member,~IEEE}

\thanks{T. Zhang, Z. Han, and Q. Ye are with the School
of Electronics, Electrical and Communication Engineering, University of Chinese Academy of Sciences, Huairou, Beijing, 108408 China. E-mail: (zhangtianliang17@mails.ucas.ac.cn, hanzhj@ucas.ac.cn, qxye@ucas.ac.cn). Huijuan Xu is with the department of Computer Science, University of California at Berkeley. B. Zhang is with the Department of Automation, BeiHang University, China. Q. Ye is the corresponding author.}
}

\markboth{IEEE Transactions On Intelligent Transportation Systems}%
{Shell \MakeLowercase{\textit{et al.}}: Bare Demo of IEEEtran.cls for IEEE Journals}
%


\maketitle

\begin{abstract}
Pedestrian detection in the wild remains a challenging problem especially when the scene contains significant occlusion and/or low resolution of the pedestrians to be detected. Existing methods are unable to adapt to these difficult cases while maintaining acceptable performance. In this paper we propose a novel feature learning model, referred to as CircleNet, to achieve feature adaptation by mimicking the process humans looking at low resolution and occluded objects: focusing on it again, at a finer scale, if the object can not be identified clearly for the first time. CircleNet is implemented as a set of feature pyramids and uses weight sharing path augmentation for better feature fusion. It targets at reciprocating feature adaptation and iterative object detection using multiple top-down and bottom-up pathways. To take full advantage of the feature adaptation capability in CircleNet, we design an instance decomposition training strategy to focus on detecting pedestrian instances of various resolutions and different occlusion levels in each cycle. Specifically, CircleNet implements feature ensemble with the idea of hard negative boosting in an end-to-end manner. Experiments on two pedestrian detection datasets, Caltech and CityPersons, show that CircleNet improves the performance of occluded and low-resolution pedestrians with significant margins while maintaining good performance on normal instances.
\end{abstract}


%
\IEEEpeerreviewmaketitle

%
%
%
%

\section{Introduction}

\IEEEPARstart{P}{edestrian} 
detection is an important problem in intelligent transportation with many real-world applications in intelligent surveillance systems  platforms, driver assistant systems, and autonomous vehicles \cite{YeSelf2017,DBLP:conf/cvpr/ZhangBOHS16,PersonSearch2018,Flores2018,DBLP:journals/tits/BroggiCGGJ09,Li2017}. Although extensively investigated, robust pedestrian detection at a long distance with a single low-cost camera remains unsolved.

Deep learning methods have achieved unprecedented success on visual object detection; nevertheless, they have trouble with adapting difficult pedestrian instances without sacrificing performance on normal instances~\cite{wang2017repulsion},~\cite{DBLP:conf/wacv/DuELD17},~\cite{DBLP:conf/eccv/ZhangLLH16},~\cite{DBLP:conf/iccv/BrazilYL17},~\cite{DBLP:conf/cvpr/ZhangBS17},~\cite{DBLP:conf/cvpr/MaoXJC17}. Take traffic scenes for example, systems have difficulty with instances that are occluded and/or have low-resolution objects, and with scenes in cluttered  backgrounds~\cite{wang2017repulsion}~\cite{zhang2018occluded}.

\begin{figure*}[tbp]
    \subfloat[FPN]{\label{Fig.1:subfig_a}
    \begin{minipage}{0.24\linewidth}
        \includegraphics[width=4cm]{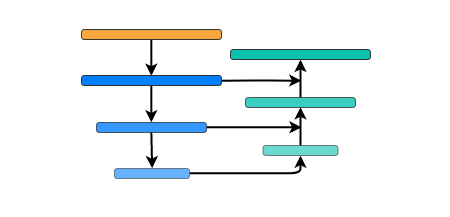}
    \end{minipage}}
    \subfloat[PANet]{\label{Fig.1:subfig_b}
    \begin{minipage}{0.24\linewidth} 
        \centering
        \includegraphics[width=4cm]{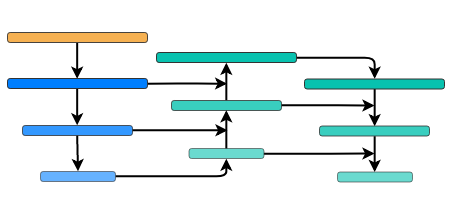}
    \end{minipage}}
    \subfloat[CircleNet (Ours)]{\label{Fig.1:subfig_c}
    \begin{minipage}{0.5\linewidth}
        \centering
        \includegraphics[width=8.2cm]{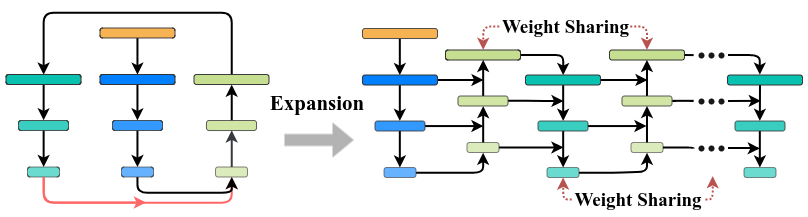}
    \end{minipage}}
    \caption{From  Feature Pyramid Network  to CircleNet. (a) Feature Pyramid Network (FPN) \cite{DBLP:conf/cvpr/LinDGHHB17}. (b) Path Aggregation Network (PANet)~\cite{liu2018path}. (c) Our proposed CircleNet is implemented as a set of feature pyramids using weight sharing path augmentation. (Lateral connection paths are illustrated in Fig. 2)} 
    \label{Fig.1}
\end{figure*}

To detect these occluded and/or low-resolution pedestrians, it is natural to consider taking higher-resolution features in the lower-level feature pyramid to assist detection using deep convolutional neural network (CNN)  \cite{DBLP:conf/cvpr/LinDGHHB17,liu2018path}.
The Feature Pyramid Network (FPN)~\cite{DBLP:conf/cvpr/LinDGHHB17}, for example, introduces an additional network branch from top to bottom to increase the feature representation capacity at lower layers (Fig.\ \ref{Fig.1:subfig_a}). 
The Path Aggregation Network (PANet)~\cite{liu2018path} adds a path augmentation from bottom to top on the basis of FPN to further boost the information flow for feature representation (Fig.\ \ref{Fig.1:subfig_b}). These approaches improve the performance of low-resolution instances; however, instances with heavy occlusion remain challenging.  

As humans, we may not be able to recognize the hard instances with only one glance, but need to focus on the specific area of the image in question. Inspired by this, we propose a feature learning framework, CircleNet, shown in Fig. 1c, which reciprocates the feature adaptation by formatting multiple top-down and bottom-up feature fusion pathways to enhance the feature representation for occluded and low-resolution objects. 
When more information pathways incorporated, we need to pay attention to the model capacity for these additional layers due to the fact that more parameters increase the chance of over-fitting. We propose using a weight sharing policy for these additional layers and experimentally validate that sharing the network parameters of the top-down/bottom-up pathways across repeating steps can maintain a healthy balance between model capacity and the generalization ability.

CircleNet is a general network architecture with FPN and PANet as its special cases. It naturally mimics the functionality of the cognitive phenomenon that the long-latency responses of the neurons contain many levels of forward and backward pipelines in the information processing of the early visual cortex \cite{lee2003hierarchical}. 
CircleNet achieves feature adaptation by mimicking the process humans looking at low resolution and occluded objects: focusing on it again, if the object cannot be identified clearly at the first time. More specifically, CircleNet uses a circling structure to implement feature adaptation, which means that pedestrians of various appearance are handled in different circles. The circling structure is constructed by a top-down branch and a bottom-up branch, which fuses features in a reciprocating manner and facilitates learning adaptive feature representation. 

The most relevant approach, Path Aggregation Network (PANet) \cite{liu2018path} adds one path augmentation from bottom to top on the basis of FPN to boost the information flow for feature representation. However with a single bottom-up path and without parameter sharing, PANet is still limited in learning adaptive features. Our CircleNet includes more cycles with multiple information flows, where useful information can flow on parallel circles or vertical paths in an adaptive manner, as shown in Fig.\ \ref{Fig.1:subfig_c}. 

The other relevant work, Cascade R-CNN \cite{CascadeRCNN2018}, consists of a sequence of detectors trained with increasing IoU thresholds, to be sequentially more selective against close false positives. Nevertheless, Cascade R-CNN requires more learnable parameters, which take the risk of over-fitting given limited training samples. In contrast, the proposed CircleNet shares the network parameters across repeating steps, and thereby maintains a healthy balance between model capacity and the generalization ability.

To take advantage of the unique feature representation in each cycle and boost the feature adaptation capability, we further propose using instance decomposition during the training and test phases. This idea is motivated by  the preliminary observations that shallow layers benefit small-scale object detection while deep cycles benefit occluded object detection. In the instance decomposition procedure, training samples are partitioned into sub-sets according to their resolution and occlusion rates, and are assigned to various circles and feature maps. As cycling goes on, Fig. 1c, we obtain adaptive feature representations corresponding to the sample distributions of various instance decomposition groups, and the network can be viewed as an ensemble classifier. Each branch will focus on detecting objects of a specific scale or occlusion ratio, and the CircleNet shares the backbone network and operates in an ensemble mode with a boosting strategy. 

The contributions of the paper are as follows:

1. A novel feature learning architecture, referred to as CircleNet, which fuses the features in a reciprocating manner and improves the capability and  adaptability of feature representation.

2. An instance decomposition strategy designed for CircleNet, which makes full use of adapted and unique feature representations at each cycle for low-resolution and occluded pedestrian detection.

3. The demonstration of state-of-the-art performance for occluded and/or low-resolution pedestrians on two standard benchmarks.

The remainder of the paper is organized as follows. Related works are reviewed in Section II. The CircleNet architecture is described in Section III and pedestrian detection with CircleNet is presented in Section IV. A discussion about CircleNet is given in Section V. Experiments are presented in Section VI and conclusions are given in Section VII.

\section{Related Work}

In the section, we first review the commonly used pedestrian detection methods. We then mainly review CNN-based pedestrian detection approaches and analyzed recent methods dealing with low-resolution and occluded issues.

\textbf{Pedestrian detection:} 
As one of the most import object sensing tasks, pedestrian detection has been extensively investigated during the past decades. Various sensors including 3-D Range Sensors~\cite{Range2016}, Near-Infrared Cameras~\cite{LeeCFH15}, Stereo Cameras~\cite{Stereo2009}, CCD Cameras~\cite{Qiao2008} and a combination of them~\cite{KrotoskyT07} have been employed. 
For visual pedestrian detection, various hard-crafted and learning-based visual features including Histogram of Gradients (HOG)~\cite{DBLP:conf/cvpr/DalalT05, DBLP:journals/pami/FelzenszwalbGMR10}, Local Binary Patterns (LBP)~\cite{DBLP:journals/pami/OjalaPM02}, Aggregated Channel Features~\cite{DBLP:journals/pami/DollarABP14}~\cite{DBLP:conf/icassp/KeZWYJ15}, Informed Haar-like features \cite{InforHaar2014,TowardsHuman18}, Rectangle Feature~\cite{Rectangle2015}, and Convolutional Neural Network (CNN)~\cite{XiaoGang2014} have been been explored. With the visual feature presentations, SVM~\cite{YeHJL13}, Random Forest~\cite{MarinVLAL13}, and cascaded models~\cite{DBLP:journals/tits/KwakKN17,Qiao2008} were used as classifiers. 

As a major branch of visual pedestrian detection method, part-based model was widely explored by first dividing a pedestrian object into parts and then training part-based models to detect pedestrians of various postures~\cite{MGTBM2013,XuVLMP14,PedersoliGHR14,LiuYDCYZ15}. Integrated with deep learning features, part-based models have shown great advantages to detect mult-view and multi-posture pedestrians, but remained challenging for pedestrians objects with low-resolution and heavy occlusion.

Another branch of pedestrian detection method rooted in feature/classifier ensemble. With tree classifier ensemble, Xu \textit{et al.} \cite{XuCQ11} targeted at achieving not only high detection accuracy but also high detection speed. With error correcting output code classification of manifold sub-classes, Ye \textit{et al.} \cite{YeHJL13} can robustly detect multi-view and multi-posture pedestrians. In \cite{BaekKK17}, the cascade implementation of the additive KSVM (AKSVM) was proposed for the application of pedestrian detection. AKSVM avoided kernel expansion by using look-up tables, and it was implemented in cascade form, thereby speeding up pedestrian detection.

\textbf{CNN-based Pedestrian Detection:} 
Early CNN-based pedestrian detection \cite{DBLP:conf/cvpr/ZhangBOHS16, DBLP:conf/cvpr/HosangOBS15} was primarily based on the RCNN structure~\cite{DBLP:conf/cvpr/GirshickDDM14,XiaoGang2014}, which relies on high-quality object proposals to achieve pedestrian localization and detection. More recently the Faster R-CNN \cite{DBLP:conf/nips/RenHGS15} architecture has become popular as it integrates region proposals with object classification for end-to-end learning. In \cite{DBLP:conf/bmvc/LiLLL18} and  \cite{object-detection-with-AP-loss}, Tiny-DSOD and AP-loss approaches were proposed to reduce the computational resource while maintaining the accuracy for detection. AP-Loss can also alleviate the extreme foreground-background class imbalance issue caused by the large number of anchors.

By borrowing general object detection frameworks to tackle pedestrian detection, these approaches have already achieved unprecedented performance. Nevertheless, detecting low-resolution and occluded pedestrians remains an open and challenging problem, as indicated by the low performance of existing state-of-the-art approaches (the miss rate is often higher than 20\% when false positive rate per image is 0.01 \cite{zhang2018occluded}).

\textbf{Low-resolution Pedestrian Detection:}
To handle low-resolution objects, a number of approaches have explored using hierarchical features in CNNs, such as SSD~\cite{DBLP:conf/eccv/LiuAESRFB16},  MS-CNN \cite{DBLP:conf/eccv/CaiFFV16}, SA-FastRCNN \cite{DBLP:journals/tmm/LiLSXFY18}, and FPN~\cite{DBLP:conf/cvpr/LinDGHHB17}. These methods leveraged fused hierarchical features to aggregate the receptive fields and representative capability from different layers. 

Semantic segmentation  \cite{DBLP:conf/iccv/BrazilYL17}, temporal features, and depth information \cite{DBLP:conf/cvpr/MaoXJC17} have also been explored to address the problem of low-resolution. Adaptive Faster R-CNN \cite{DBLP:conf/cvpr/ZhangBS17} was used to process low-resolution instances by optimizing the scales and aspect ratios of region proposals and the resolution of the image. A cascaded Boosted Forest (BF) was applied on up-sampled feature maps to detect low-resolution pedestrians  \cite{DBLP:conf/eccv/ZhangLLH16}. In \cite{SmallScaleEccv2018-Graininess}, fine-grained information was incorporated into features to make them more discriminative for human body parts. In \cite{SmallScaleEccv2018-T}, topological line localization (TLL) and temporal feature aggregation were used to detect low-resolution pedestrians.

\begin{figure}[tbp]
    \centering
    \includegraphics[width=1.0\linewidth]{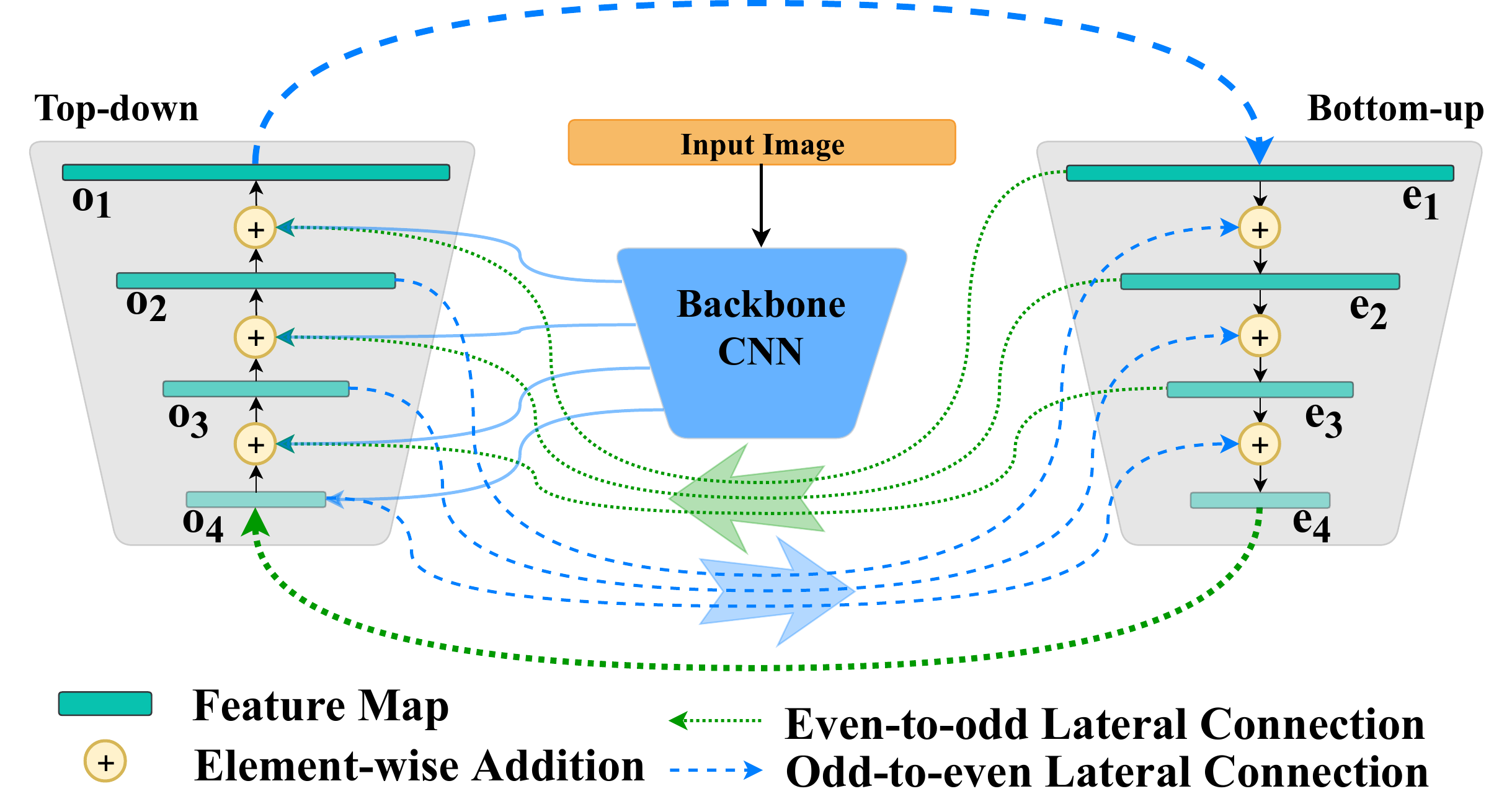}
    \caption{\label{Fig.arc} CircleNet architecture. CircleNet is made up of a backbone network and  top-down and bottom-up branches. The top-down branches enforce semantics by up-sampling and concatenating deep-layer features. The bottom-up branches enlarge the receptive field and incorporate context information by down-sampling and concatenating features from shallow layers. The top-down and bottom-up branches construct a circling network that fuses  features in a reciprocating manner. }
    \label{fig_arc}
\end{figure}

\textbf{Occlusion Handling:} One effect of occlusion is that it significantly aggravates the appearance of pedestrian instances. To address this issue, one simple strategy is to use a classifier ensemble or detector \cite{DBLP:conf/cvpr/OuyangW12}\cite{DBLP:conf/iccv/TianLWT15}. In \cite{CascadeRCNN2018}, the Cascade R-CNN was proposed to address these problems. It consists of a sequence of detectors trained with increasing the Intersection over Union (IoU) thresholds, to sequentially select difficult instances.

Another effect of occlusion is that it causes the loss of certain pedestrian parts and significantly increases the difficulty of spatial localization. To address the problem, 
a pedestrian object was decomposed into parts, which can be used to model the partial occlusion of objects and the attention of classifiers and features. Wang \textit{et al.} \cite{DBLP:conf/iccv/TianLWT15} proposed the DeepPart model where each part detector is a strong detector that can detect pedestrian by observing only a part of a proposal. In \cite{wang2017repulsion} a repulsion loss (RepLoss) approach was designed to enforce spatial localization in crowd scenes. With RepLoss, each proposal is forced to be close to its designated target, and keep it away from other ground-truth objects. In \cite{zhang2018occluded}, the Faster R-CNN with attention guidance (FasterRCNN-ATT) was proposed to detect occluded instances. Assuming that each occlusion pattern can be formulated as a combination of body parts, a part attention mechanism was used to represent various occlusion patterns in one single model.

Existing approaches have included effective strategies to process low-resolution and occluded instances, but often require empirically defined strategies. Some of them achieve higher performance on difficult instances but get lower performance on normal instances. In this paper we propose to mimic the human cognition processes to handle these difficult instances without sacrificing the performance on normal instances. The proposed CircleNet leverages a recurrent structure with parameter sharing to realize feature adaptation. The top-down and bottom-up branches in multiple circles facilitate fusing features in a cognitively plausible way. In contrast, PANet~\cite{liu2018path} adds only one bottom-up path and do not use parameter sharing.

\textbf{Hard Instance Handling:} OHEM \cite{DBLP:conf/cvpr/ShrivastavaGG16} is a widely used to handle the hard instances. It mines hard examples according to the training loss of instances and makes the learned models more discriminative. Our instance decomposition strategy is different from OHEM. It is based on the heuristics that the normal instances and hard instances have different feature representations. An instance decomposition strategy is used with the deep circles paying more attention to hard examples while the shallow circles paying more attention to normal instances. In contrast, OHEM uses the same features for all hard and normal instances. 

With larger receptive fields, deep circles are easy to detect occluded pedestrians than shallow ones. Existing works \cite{DBLP:conf/bmvc/Wang17, DBLP:conf/bmvc/ZagoruykoLLPGCD16, DBLP:conf/iccv/GidarisK15} have explored context information for object detection, and find that appropriate context is helpful for detecting objects with occlusion. We take advantage of more context information to learn features that are adaptive to hard pedestrian examples. Considering the property of context information and the circling architecture, we argue that the proposed instance decomposition strategy is effective in the CircleNet structure.

\section{CircleNet}

In this section, we first introduce the architecture of CircleNet. We then describe the reciprocating feature adaptation and instance decomposition performed with CircleNet.

\subsection{Architecture}

CircleNet is made up of a backbone network and a circling architecture (Fig.\ \ref{Fig.arc}). The top-down branches enforce semantics by up-sampling and concatenating deep-layer features. The bottom-up branches enlarge the receptive field and incorporate context information by down-sampling and concatenating features from shallow layers. The top-down and bottom-up branches construct a circling network that fuses  features in a reciprocating manner. The backbone network is a commonly used fully convolutional network, which computes a feature map hierarchy consisting several scales with a scaling factor of 2.

For the top-down (odd) pathway, let $o_{n}, n=1,...,N$, denote the $n^{th}$ feature map, and $o_{n+1}$ and $e_n$ denote the input and the side-output, respectively. $e_n$ stores the information from the current layer and $o_{n+1}$ brings more high level information from the deep layers. Similarly, for the bottom-up (even) pathway, let $e_{n+1}$ denote the $(n+1)^{th}$ feature map, and $e_{n}$ and $o_{n+1}$ denote the input and the side-output, respectively. $o_{n+1}$ stores the information from the current layer and $e_{n}$ is used to generate new features. Cascading the multiple top-down and bottom-up pathways together 
is implemented as a cascaded feature pyramid network.

To upgrade the cascading architecture to a circling one, we share weights for all top-down pathways and for all bottom-up pathways from different circles, as shown in  Fig.\ \ref{Fig.1:subfig_c}. 
In this way, we keep a minimal number of learnable parameters and learn features that are adaptive to different instances. This means we only need to learn weights for an odd branch and an even pathway $\{w^e, w^o\}$. By weight sharing, we update the cascading architecture to a circling architecture, which learns different features in each circle $t$ as:
\begin{equation}
    \mathbf{x}_t = F(\mathbf{x}_{t-1}, w^e, w^o),
\end{equation}
where $\mathbf{x}_t = \{o_n^t, e_n^t\}$, $w^e = \{w^e_n\}$, and $w^o = \{w^o_n\}$, $n=1,..., N, t=1,...,T$. $n$ is the index of the feature maps and $t$ is the index of circles (see Section \ref{ReciprocatingFeatureAdaptation} for details). $F(\cdot)$ denotes the feature extraction function.

\subsection{Reciprocating Feature Adaptation}
\label{ReciprocatingFeatureAdaptation}
The circling architecture implements feature fusion and adaptation in a reciprocating manner by concatenating multiple top-down and bottom-up pathways. Each pathway uses lateral connections to fuse features which come from the backbone network or a pathway. 
The top-down pathway follows the implementation of FPN \cite{DBLP:conf/cvpr/LinDGHHB17}. In the feed-forward procedure of any circle, the feature of the $n^{th}$ top-down layer at $t^{th}$ circle, $o_{n}^t$, is generated by fusing the features from the backbone CNN and the top-down layer $o_{n+1}$, or features from the bottom-up layer $e_n^t$ and $o_{n+1}$, as
\begin{equation}
\begin{split}
o_n^t &= F_n(e_n^t, o_{n+1}^t,w_n^{e}) \\
      &= w^{e_{11}}_{n} \ast (w^{e_{33}}_{n} \ast e_n^t) + \uparrow o_{n+1}^t,
\label{eq_deep2shallow}
\end{split}
\end{equation}
where $w_n^{e} = \{w_n^{e_{11}}$, $w_n^{e_{33}}\}$ are $1 \times 1$ and $3 \times 3$ convolutional filters to fuse features and generate $o_{n}^t$. $\uparrow$ denotes the up-sampling operation.

\begin{figure}[tbp]
    \centering
    \includegraphics[width=1\linewidth]{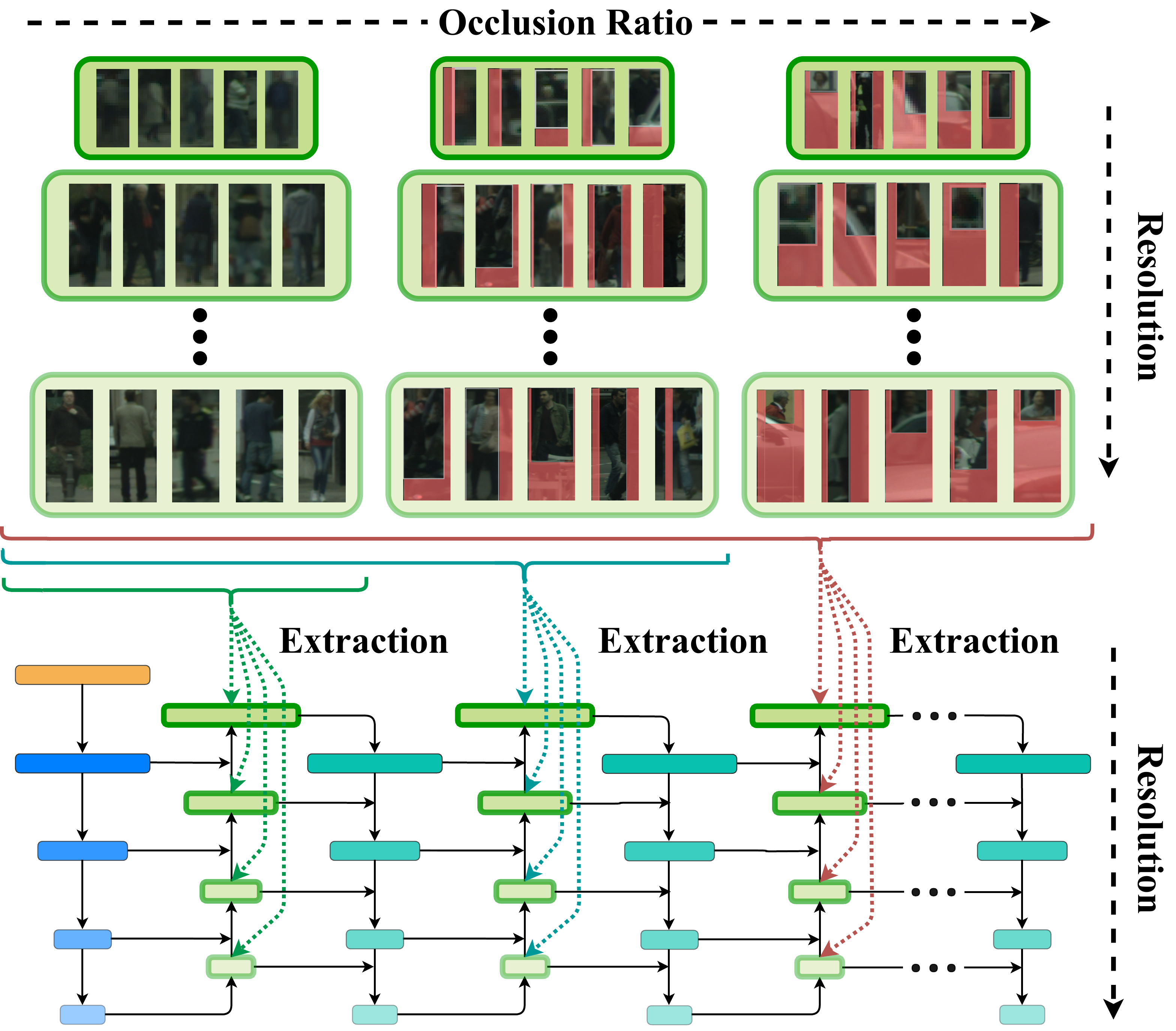}
    \caption{\label{Fig.instance_decomposition} An illustration of instance decomposition during the training phase. (Best viewed in color and zoomed in) }
\end{figure}

Similarly, the features of the $(n+1)^{th}$ bottom-up layer, $e_{n+1}^t$, are generated by fusing the features from the bottom-up layer $o_{n+1}^{t-1}$ and its previous bottom-up layer $e_{n}^t$, as
\begin{equation}
\begin{split}
e_{n+1}^t &=F_n(e_n^t, o_{n+1}^{t-1},w_{n+1}^{o}) \\
          &=w^{o_{11}}_{n+1} \ast ( w^{o_{33}}_{n+1} \ast o_{n+1}^{t-1}) + \downarrow e_{n}^t,
\label{eq_shallow2deep}
\end{split}
\end{equation}
where $w_n^o = \{w_n^{o_{11}}$, $w_n^{o_{33}}\}$ are $1 \times 1$ and $3 \times 3$ convolutional filters to fuse features and generate $e_{n+1}$. 
$\downarrow$ denotes the down-sampling operation.

The up-sampling operation is implemented by the nearest neighbor interpolation \cite{ni2008adaptable}. Nearest-neighbor interpolation (also known as proximal interpolation or, in some contexts, point sampling) is a simple method of multivariate interpolation in one or more dimensions. The nearest neighbor interpolation selects the value of the nearest point which does not consider the values of neighboring points, yielding a piecewise-constant interpolant. The down-sampling operation is implemented by a strided convolution.

In the feed-forward procedure, the top-down pathways hallucinate higher resolution features by up-sampling spatially coarser but semantically stronger feature maps from higher pyramid levels. These features are then enhanced with features from either the backbone network or the bottom-up pathway, via lateral connections. Each lateral connection merges feature maps of the same spatial size from the bottom-up pathway and the top-down pathway. The top-down pathways further fuse features by reducing their resolution but enforcing their semantics. Reducing the resolution of feature maps can enlarge the receptive field and collect context information, which is crucial to detect low-resolution and occluded instances. 

By using multiple top-down and bottom-up pathways as shown in Eqs.\ \ref{eq_deep2shallow} and  \ref{eq_shallow2deep}, the CircleNet implements reciprocating feature fusion that provides a higher probability of producing features with both strong semantics and context information.

\subsection{Instance Decomposition}
\label{InstanceDecomposition}

With reciprocating feature fusion, the circling architecture has the potential to learn features that are adaptive to pedestrian instances of various appearance. This motivates us to empirically decompose the instances into different feature maps on the circles, so that normal and hard instances are handled with proper features. In experiments, we observed that deep circles pay more attention to samples with occlusion, while for samples with no occlusion the detectors perform well in the shallow circles. Thereby, an instance decomposition strategy is proposed so that difficult instances are decomposed into deep circles according to the training loss or the occlusion ratio. Low-resolution instances are decomposed into shallow feature maps.
 
Empirically, we use circle-based and layer-based instance decomposition and partition the training set $D$ into $N\times T$ sub-sets, where $N$ and $T$ are the number of feature maps and circles, as shown in Fig.\ \ref{Fig.instance_decomposition}.

Along the circles, $D$ is decomposed into sub-sets with some instance overlapping, as given by
\begin{equation}
D = \mathop{\cup} \limits_{t} D^t,  D^t\cap D^{(t+1)} \neq \emptyset,
\label{eq_InstanceDecomposition}
\end{equation}
where $D^t = \sum\limits_{n} D_n^t $. $D^t$ is a sub-set of easy instances, while $D^{(t+1)}$ is a larger sub-set which includes additional instances that are harder than $D^t$. The harder instances are identified by their occlusion rates or training loss.

Along the feature maps in each circle, $D^t$ is decomposed into sub-sets without instance overlap, as

\begin{equation}
D^t = \mathop{\cup} \limits_n {D_n^t}, D_n^t \cap D_{n+1}^t=\emptyset,
\end{equation}
where $D_{n+1}^t$ is a sub-set of instances whose resolutions are higher than $D_n^t$. We assign high resolution instances to deep layers and low resolution instances to shallow layers.

\section{Pedestrian Detection with CircleNet}
Using the CircleNet as a backbone network, we implemented pedestrian detection by using the region proposal network (RPN) to generate object proposals at each feature layer. The CircleNet is optimized using stochastic gradient descent (SGD) in an end-to-end manner.

\subsection{Implementation}
CircleNet contains a backbone CNN for basic feature extraction and a circling module for feature adaptation. The basic feature maps for ResNets \cite{DBLP:conf/cvpr/HeZRS16} are the outputs of last residual blocks for conv2, conv3, conv4, and conv5. These basic feature maps are fed to the circling module to generate adapted feature maps $\{o^t_n, e^t_n\}, n=1,\dots,4, t=1,\dots,T$, which are used for region proposal generation, object classification, and bounding box regression. 

An RPN is a sliding-window class-agnostic object detector for generating region proposals, and a predictor classifies each proposal and refines the proposal for object localization. We use the RPN and the predictor on each circle and note that the parameters of the RPNs and the predictor heads are shared across all circles. The procedure of detection involves several steps. First the input images go through the backbone CNN and the circling module, and the RPNs are used to generate proposals. Second the features of each proposal are generated by cropping from the adapted features, \textit{e.g.}, $\{o^t_n\}$. More specifically, the proposal features are extracted from the $n^{th}$ layer of feature pyramid according to the proposal' size in the current circle ($T=t$), where $n = \lfloor k_0 + log_2(area/224) \times \theta \rfloor$, $\theta$ is a hyper-parameter relevant to the dataset. ``area" denotes the area of the proposal. In experiment, we set $k_0=4$ and $\theta=log_2(2)$, which means that the features of the small proposals are extracted from high-resolution pyramid layers while the features of the large proposals are extracted from low-resolution pyramid layers. Third, a Region-of-Interest (RoI) pooling layer produces the same length of feature for each proposal. Finally, these features are fed to the predictors to determine the detection.

\subsection{Learning}

\begin{figure}[t]
    \centering
    \includegraphics[width=0.9\linewidth]{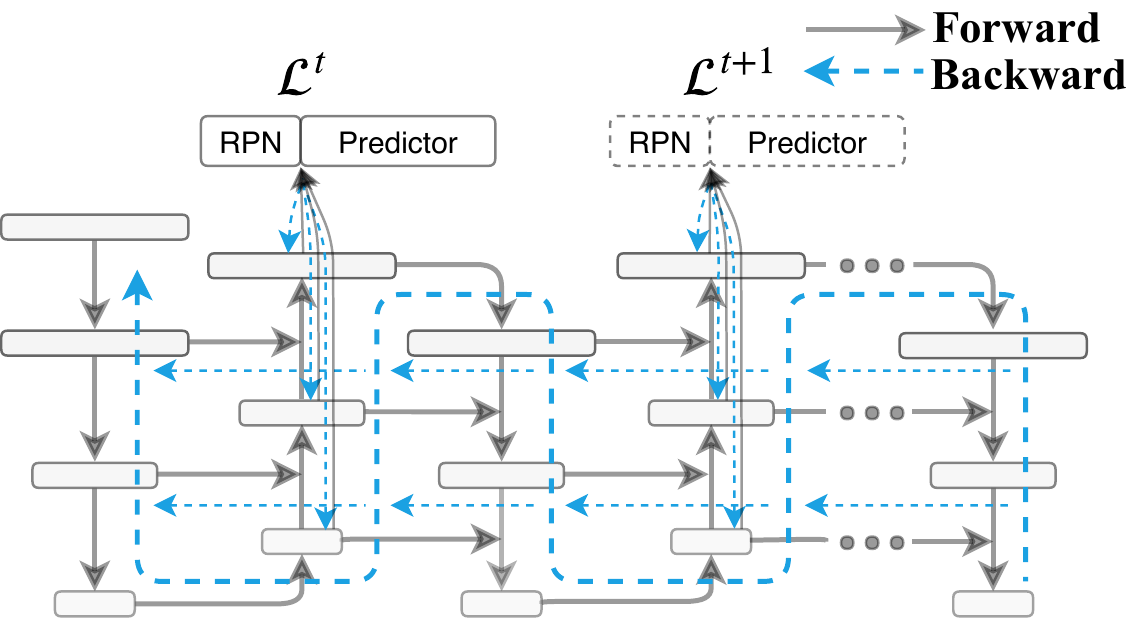}
    \caption{\label{Fig.Optimization} On each of the feature layer of the top-down pathway, an RPN and a predictor is implemented for pedestrian detection. During the learning procedure, forward connections (solid lines with arrows) along the CircleNet are for feature extraction and object score prediction. Backward connections (dashed lines with arrows) are for gradient propagation.}
\end{figure}

We train the network by optimizing both the classification and regression tasks for the RPNs and the predictors. The softmax loss function is used for classification, and the smooth L1 \cite{Girshick15} loss function is used for regression. The layers on a top-down pathway focus on aggregating the semantic information from deep layers. We therefore add a pseudo segmentation loss\footnote{A pseudo-mask is generated for each image by setting the pixels in the pedestrian bounding boxes to 1 and in the background to 0.} on the top-down pathways. By introducing a segmentation layer which drives the features to focus on pedestrian regions and to suppress the negative samples from cluttered backgrounds. The loss function, defined as
\begin{equation}
\begin{split}
    \mathcal{L} = \sum_t \mathcal{L}^{t} = \sum_t(\sum_n({\mathcal{L}\scriptstyle{rpn\_cls}}^{t}_{n} + {\mathcal{L}\scriptstyle{rpn\_reg}}^{t}_{n} + \\ {\mathcal{L}\scriptstyle{cls}}^{t}_{n}+{\mathcal{L}\scriptstyle{reg}}^{t}_{n})+{\mathcal{L}\scriptstyle{seg}}^{t}),
\end{split}
\end{equation}
is applied to each RPN, predictor and segmentation layer. The softmax loss ${\mathcal{L}\scriptstyle{rpn\_cls}}$ and ${\mathcal{L}\scriptstyle{cls}}$ optimize the classes for anchors and proposals. The regression loss $\mathcal{L}\scriptstyle{rpn\_reg}$ optimizes the relative displacement between anchors and ground truth, and ${\mathcal{L}\scriptstyle{reg}}$ optimizes the relative displacement between proposals and ground truth. 
The forward and backward propagation procedures are shown in Fig.\ \ref{Fig.Optimization}, which shows that the feature/gradient can be directly propagated along circles.

\section{Understanding CircleNet}

From the perspective of learning, CircleNet implements a special kind of classifier ensemble. For multiple feature layers on the circles, we have multiple classifiers, each of which is responsible for a sub-set of instances. In the deep learning framework, the hard instances have large training loss, which means that they have large weight during the learning procedure. This is actually a boosting-like learning strategy. In the training phrase, the pedestrian instances of different appearance, \textit{e.g.}, resolution and occlusion rates, are processed with multiple base classifiers. In the test phase, the maximum classification scores are used to determine final detections. 

By sharing the backbone based in a feature reciprocating procedure, we can actually boost the base classifiers learned on different sub-sets. As a result, the feature (base classifier) achieved on difficult samples can also be adapted to normal samples. This not only benefits pedestrian detection but also leads to a new ensemble method in the deep learning framework.

\section{Experiments}

In this section, the experimental protocols are introduced and  the effects of the CircleNet are analyzed. The performance of pedestrian detection with CircleNet and comparisons with other state-of-the-art detectors are presented. Ablation experiments are carried out to validate the effectiveness of CircleNet backbone for pedestrian detection in challenging traffic scenes.

\subsection{Experimental Protocols}

\textbf{Datasets:} Two of the most popular datasets for pedestrian detection, \textbf{Caltech}~\cite{DBLP:conf/cvpr/DollarWSP09} and \textbf{CityPersons}\cite{DBLP:conf/cvpr/ZhangBS17}, are used to evaluate CircleNet. The Caltech dataset contains approximately 10 hours of street-view video taken with a camera mounted on a vehicle. The most challenging aspect of the dataset is the large number of low-resolution pedestrians. We sample 42,782 images from set00 to set05 for training and 4,024 images from set06-set10 for testing. 

The CityPersons dataset is built on the semantic segmentation dataset Cityscapes \cite{DBLP:conf/cvpr/CordtsORREBFRS16}. This dataset contains 5,000 images (2,975 for training, 500 for validation, and 1,525 for testing) captured from 18 cities in Germany at three different seasons and various weather conditions. The scene in CityPersons dataset is much more ``crowded" than Caltech dataset, and the challenging aspects of the pedestrian objects include complex backgrounds, low resolution, and heavy occlusion.

\textbf{Evaluation Metrics:}
The experiments are conducted on instances across different occlusion levels, including:
(1) \textbf{Reasonable:} visibility $\in$ $[0.65, \infty]$ \& {Height$\geq$50 } (pixels);
(2) \textbf{None:} visibility $\in$ $[1.0, \infty]$;
(3) \textbf{Partial:} visibility $\in$ $[0.65, 1.0)$;
(4) \textbf{Heavy:} visibility $\in$ $[0.2, 0.65)$;
and normal and low-resolution instances, as:
(1) \textbf{Height$\geq$50} (pixels);
(2) \textbf{Height$\geq$20} (pixels).

In most previous pedestrian detection works, the miss rate (MR) over false positive per image (FPPI) is commonly used as the evaluation protocol \cite{DBLP:conf/cvpr/ZhangBS17}\cite{DBLP:conf/cvpr/DollarWSP09}. This is preferred to precision recall curves for certain tasks, \textit{e.g.}, automotive applications, as typically there is an upper limit on the acceptable false positives per image rate independent of pedestrian density. Such a protocol produces miss rates against FPPI by varying the threshold on detection scores. To emphasize the importance of missing detections, a log-average miss rate is used to summarize pedestrian detector performance, by averaging miss rates at nine FPPI rates in the range of [$10^{-1}, 10^{0}$].

A detected box is classified as a true positive, if its classification score is larger than a threshold when the box is matched with a ground-truth box, \textit{i.e.}, their IoU exceeds 50 percent. Each bounding box and ground truth can be matched at most once. Detections with the highest confidence are matched first; if a detected bounding box matches multiple ground-truth bounding boxes, the match with the highest overlap is used. Unmatched detected bounding boxes are classified as false positives.

\textbf{Implementation Details:}
FPN~\cite{DBLP:conf/cvpr/LinDGHHB17} with the ResNet50 backbone~\cite{DBLP:conf/cvpr/HeZRS16} is used as the baseline detector. For a fair comparison with other state-of-the-art methods, we up-sample the image resolution to $900 \times 1200$, fine-tune the network trained on CityPersons. On the CityPersons, we follow the settings in \cite{DBLP:conf/eccv/ZhangWBLL18} and up-sample the images with a 1.3$\times$ ratio and fine-tune the network pre-trained on ImageNet \footnote{Some recent works fine-tune the network pre-trained on Citypersons to detection pedestrians on Caltech. This can aggregates the performance at the cost of training complexity.}.

\begin{figure}[tbp]
    \subfloat[]{\label{fig:feature_adaptation_2:subfig_a}
    \begin{minipage}{1\linewidth}
        \centering
        \includegraphics[width=8.4cm]{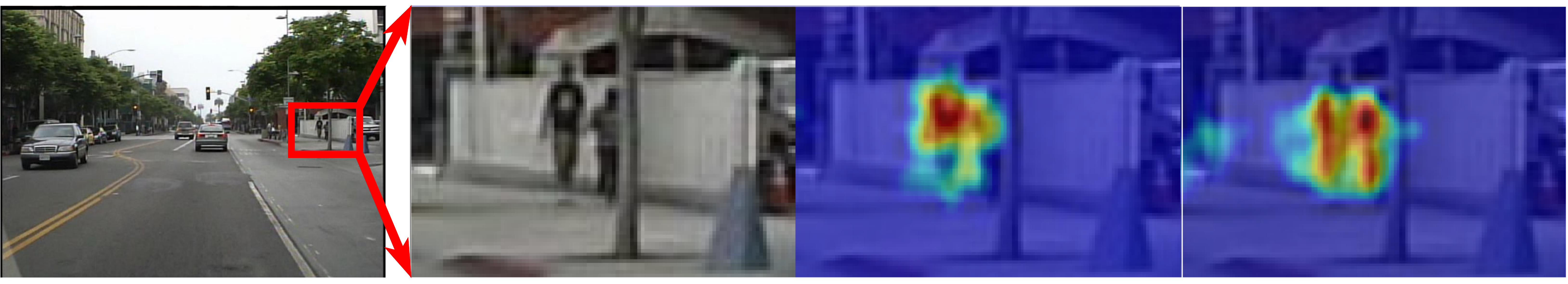}
    \end{minipage}}

    \subfloat[]{\label{fig:feature_adaptation_2:subfig_b}
    \begin{minipage}{1\linewidth}
        \centering
        \includegraphics[width=8.4cm]{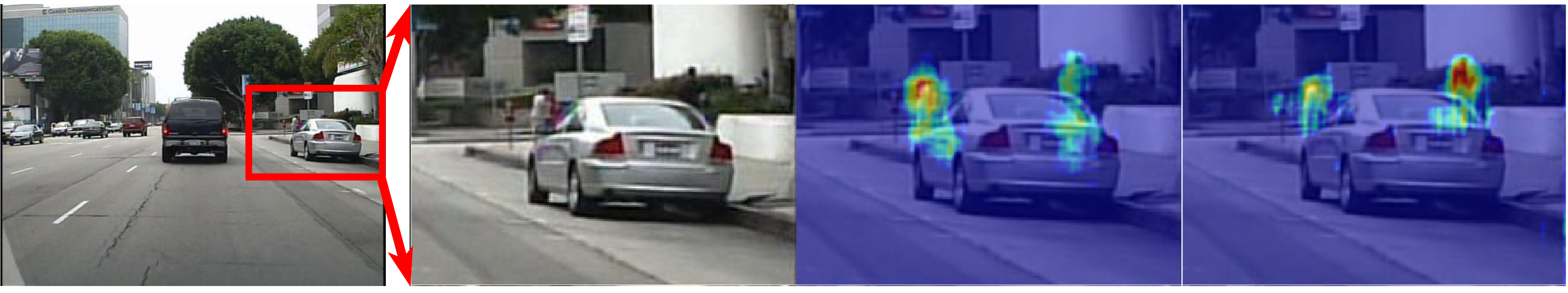}
    \end{minipage}}
    
    \subfloat[]{\label{fig:feature_adaptation_2:subfig_c}
    \begin{minipage}{1\linewidth}
        \centering
        \includegraphics[width=8.4cm]{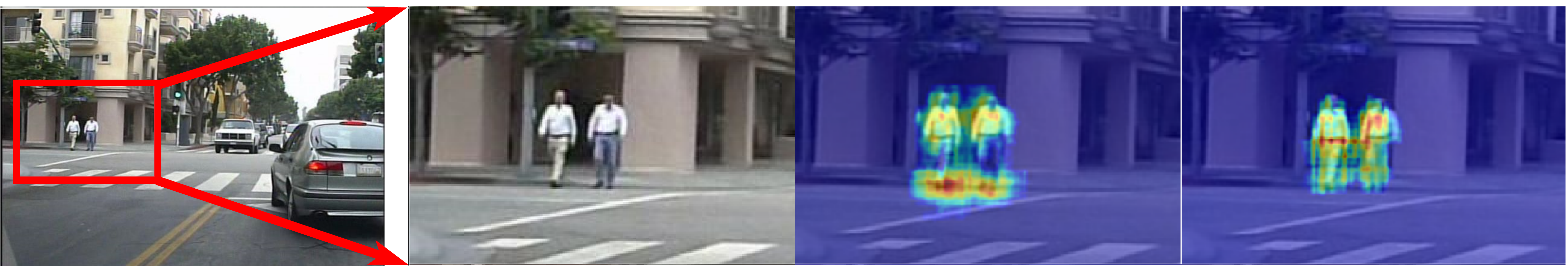}
    \end{minipage}}
    \caption{\label{Fig.feature_adaptation_2} Visualization of feature adaptation. The first column shows the original image. A cropped patch with a pedestrian is presented in the second column. The visualization of the feature map that from Circle-1 can be seen in the third column and the fourth column shows the Circle-2. (Best view in color)}
    \vspace{-0.6cm}
\end{figure}

\subsection{Ablation Studies}

\begin{table*}[]
\centering
\fontsize{7}{9}\selectfont
\begin{tabular}{c|c|ccccc|cccc}
\hline
\multirow{2}{*}{Model} & \multirow{2}{*}{Description} & \multicolumn{5}{c|}{Height$\geq${}50}                                                                     & \multicolumn{4}{c}{Height$\geq${}20}                                              \\ \cline{3-11}
                      &                           & All                  & None                 & Partial              & Heavy                & Reasonable           & All                  & None                 & Partial              & Heavy                \\ \hline \hline
FPN \cite{DBLP:conf/cvpr/LinDGHHB17}   & Baseline                  & 31.21                & 12.72                & 37.40                 & 82.96                & 15.79                & 59.85                & 48.39                 & 69.41                & 89.71                \\ \hline
CircleNet-1/2          & PANet structure               & 30.02                & 12.76                & 37.82                & 79.20                 & 15.80                 & 57.26                & 45.48                & 68.98                & 88.72                \\
CircleNet-1            & One circle ($T=1$)                & 28.74                & {\ul \textbf{10.85}} & {\ul \textbf{34.28}} & 79.87                & {\ul \textbf{13.75}} & {\ul \textbf{57.00}}    & {\ul \textbf{45.21}} & 68.29                & 88.86                \\
CircleNet-2            & Two circles ($T=2$)             & {\ul \textbf{28.12}} & 11.58                & 36.49                &75.08 & 14.63                & 57.25                & 45.62                & 69.27 & {\ul \textbf{88.19}} \\
CircleNet-3            & Three circles ($T=3$)            & 28.69                & 11.66                & 36.32                & {\ul \textbf{75.01}}                & 14.72                & 58.71                & 48.89                & {\ul \textbf{68.18}}                & 88.65                \\ \hline
FPN \cite{DBLP:conf/cvpr/LinDGHHB17} (trained with occluded instances)          & Baseline                  & 25.88                & 12.58                & 32.03                & 64.01                & 14.85                & 57.02                & 48.51                & 64.33                & 79.33                \\ \hline
CircleNet-2 (trained with occluded instances)   & Two circles               & {\ul \textbf{24.14}} & 12.84                & {\ul \textbf{28.74}} & {\ul \textbf{54.66}} & 15.02                & {\ul \textbf{55.36}} & {\ul \textbf{46.97}} & 65.77 & {\ul \textbf{75.05}} \\ \hline
\end{tabular}
\caption{Detection performance of FPN and CircleNet on the Caltech test set. $\mathrm{MR}^{-2}$ is used to compare the performance. Lower score indicates better performance.}
\label{TabCaltechAblation}
\end{table*}

\begin{figure*}[tbp]
    \centering
    \includegraphics[width=1\linewidth]{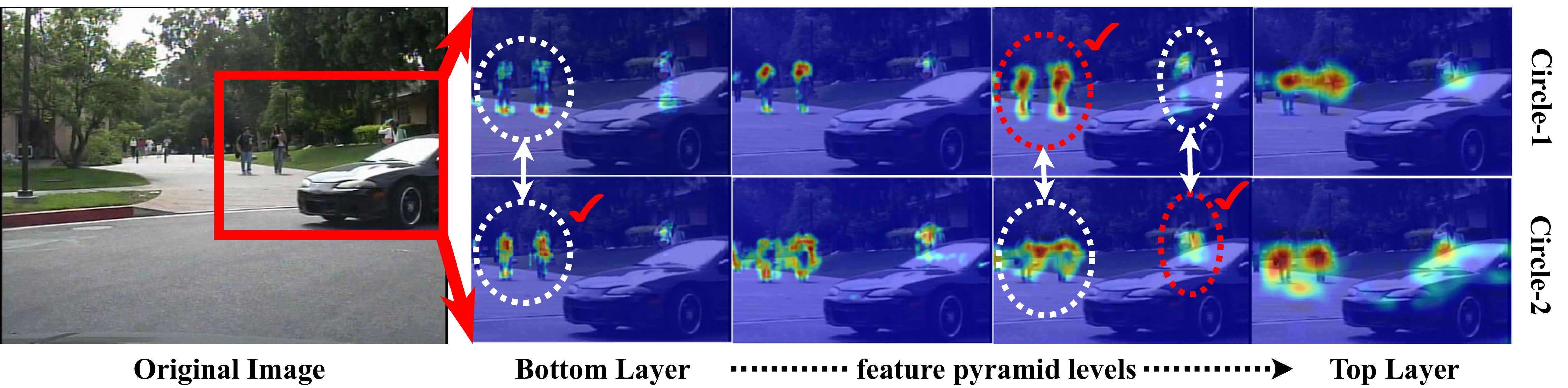}
    \caption{\label{Fig.Feature_8maps} Visualization of feature adaptation across circles. (Best view in color)}
\end{figure*}

\textbf{Circling Architecture:} 
We define the information through a bottom-up pathway and a top-down pathway as one circle, and use the FPN, which has no circling architecture, as a baseline model. The half-circling architecture (CircleNet-1/2) that has an additional bottom-up pathway to FPN is a PANet~\cite{liu2018path}. The CircleNets with one, two, and three circles are labeled as CircleNet-1 ($T=1$), CircleNet-2 ($T=2$), and CircleNet-3 ($T=3$), respectively.

In Table\ \ref{TabCaltechAblation}, it can be seen that CircleNet-1 outperforms FPN and CircleNet-1/2, while CircleNet-2 outperforms CircleNet-1. CircleNet-3 has slightly lower performance than CircleNet-2 on the ``All" (height$\geq$50) and (height$\geq$20) sub-set.
The reason could be that by increasing the number of circles, the training difficulty increases. Considering that CircleNet-3 can provide sufficient feature layers to represent pedestrians of various normal and hard instances, we do not test more circles in following experiments.

Table\ \ref{TabCaltechAblation} shows that the CircleNets outperform the baseline FPN and PANet (CircleNet-1/2) by significant margins. On the ``All" and ``Reasonable" sub-set (height$\geq$50), CircleNet-2 outperforms FPN by 3.09\% (28.12\% vs. 31.21\%) and 1.16\% (14.63\% vs. 15.79\%). On the ``None" and ``Partial" (Height$\geq$50) occlusion sub-set, CircleNet-2 outperforms FPN by 1.14\% (11.58\% vs. 12.72\%) and 0.91\% (36.49\% vs. 37.40\%), respectively. On the ``Heavy" occlusion sub-set (height$\geq$50), CircleNet outperforms FPN by 7.84\% (75.08\% vs. 82.96\%). On ``All" sub-set (Height$\geq$20), CircleNet-2 noticeably outperforms FPN by 2.60\% (57.25\% vs. 59.85\%). CircleNet-2 outperforms PANet (CircleNet-1/2) by 1.9\% (28.12\% vs. 30.02\%) on the “All” subset of Caltech dataset. For a fair comparison, the compared PANet does not use adaptive feature pooling. When training the networks with occluded instances as shown in the last two rows of Table\ \ref{TabCaltechAblation},  CircleNet still outperforms the baseline approach, particularly for low-resolution instances. 

\begin{figure}[tbp]
    \subfloat[]{\label{Fig:proposals:subfig_a}
    \begin{minipage}{0.5\linewidth}
        \centering
        \includegraphics[width=4.5cm]{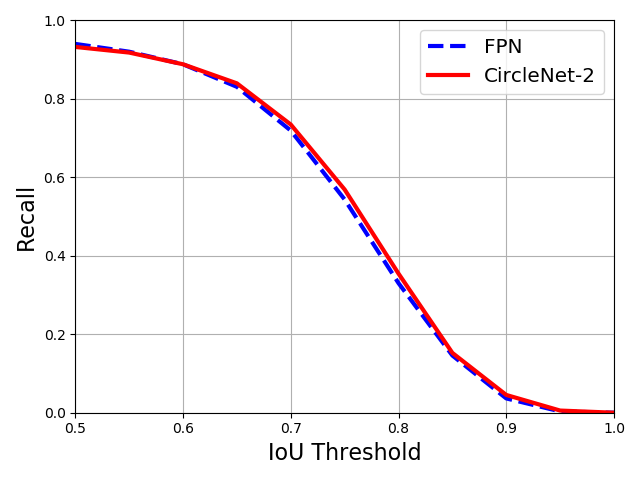}
    \end{minipage}}
    \subfloat[]{\label{Fig.proposals:subfig_b}
    \begin{minipage}{0.5\linewidth}
        \centering
        \includegraphics[width=4.5cm]{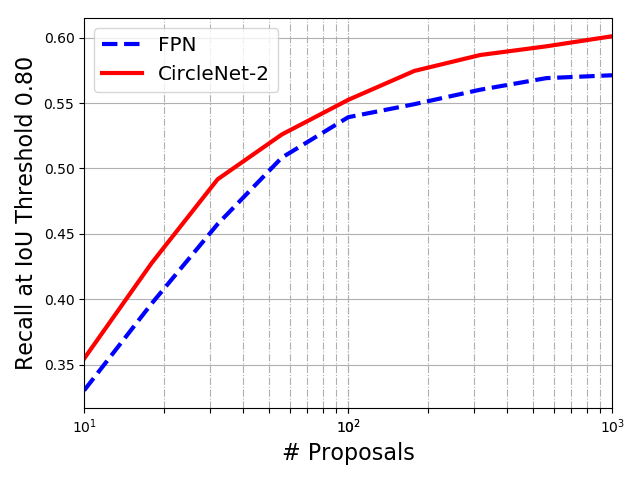}
    \end{minipage}}
    \caption{\label{Fig.proposals} Evaluation of the effect of CircleNet for region proposal network (RPN). (a) Recall rates under different IoU thresholds. (b) Recall rates under different numbers of region proposals.}
\end{figure}

The proposed CircleNet also boosts the performance of the region proposal network (RPN) under two commonly used metrics. As shown in Fig. \ref{Fig.proposals}, the first metric evaluates the recall rate under different IoU thresholds. The second metric evaluates the variation of recall rate under different numbers of region proposals. CircleNet outperforms the baseline FPN by significant margins in terms of both metrics.

As is known, down-stream features bring not only the semantic information but also noise. To alleviate the noise, we do not use the down-steam feature directly. Instead, we use a convolution layer to encode the fused features from top-down and bottom-up pathways to filter the noises from down-stream features.

\begin{table}[]
\centering
\fontsize{8}{10}\selectfont
\begin{tabular}{c|c|c|c|c}
\hline 
\textbf{Method}                                                     & \textbf{FPN} & \textbf{CircleNet-1} & \textbf{CircleNet-2} & \textbf{CircleNet-3} \\ \hline  \hline
\begin{tabular}[c]{@{}c@{}}Inference Time\\ (ms/image)\end{tabular} & 51           & 61                   & 74                   & 87                   \\ \hline 
\end{tabular}
\caption{Comparison of the detection speeds on Caltech dataset.}
\label{TableInferenceTime}
\end{table}

In Table \ref{TableInferenceTime}, we compare the inference speeds of the FPN and CircleNet. FPN takes 51ms to process an image while CircleNet-1, CircleNet-2 and CircleNet-3 take 61ms, 74ms, and 87ms, respectively. Circle-Net can significantly improve the detection accuracy with moderate computational cost overhead. With a Telsla 1080TI GPU, the best CircleNet-2 runs at 13.5 FPS, which can feed the requirement of many real-world applications.

\textbf{Feature Adaptation:}
Feature adaptation is implemented in a reciprocating manner by concatenating multiple top-down and bottom-up pathways. Fig.\ \ref{Fig.feature_adaptation_2} presents the visualization of the features from different circles (Circle-1 and Circle-2) in CircleNet-2.

\begin{figure}[tbp]
    \centering
    \includegraphics[width=1.0\linewidth]{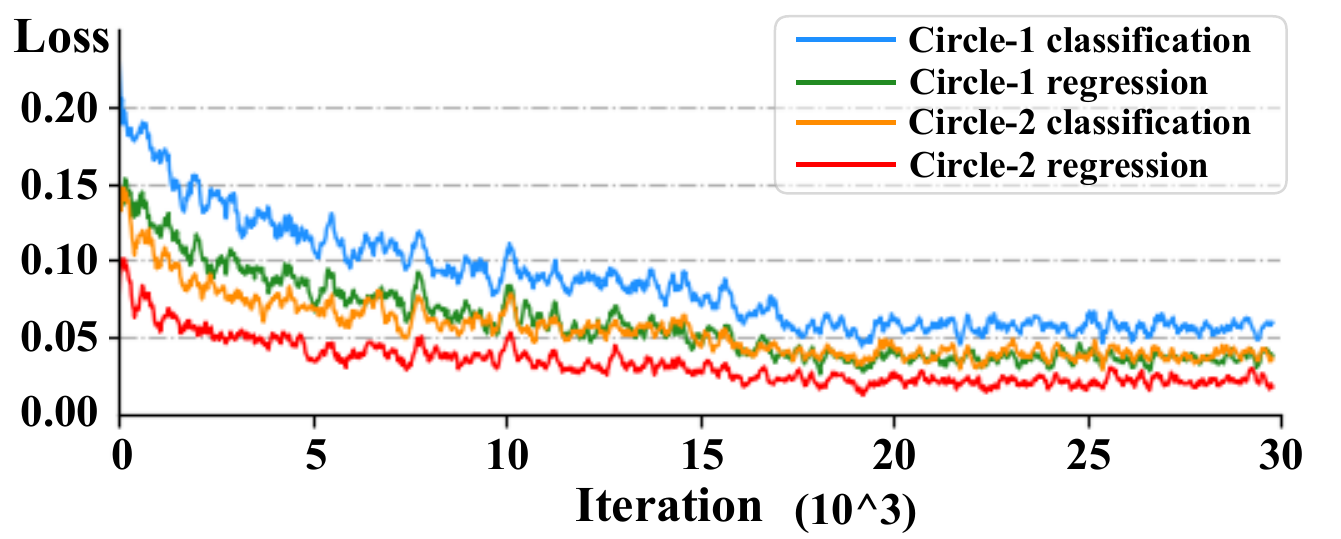}
    \caption{\label{Fig.loss} Loss comparison of Circle-1 and Circle-2. (Best view in color)}
\end{figure}

In Fig.\ \ref{fig:feature_adaptation_2:subfig_a}, a pedestrian is almost ignored by the first circle (Circle-1) as the occlusion. With feature adaptation, the second circle (Circle-2) learns representative features of the pedestrian. Fig.\ \ref{fig:feature_adaptation_2:subfig_b} shows two circles learn different features that are adaptive to different occlusion patterns. In Fig.\ \ref{fig:feature_adaptation_2:subfig_c} the learned features of the first circle activate some background areas, while those of the second circle activate the pedestrian boundary.

Fig.\ \ref{Fig.Feature_8maps} further shows the effect of feature adaption contributed by CircleNet. In the bottom layer, Circle-2 better activates pedestrian regions than Circle-1, while in the top layer, Circle-1 performs better on pedestrians with no occlusion. For pedestrians with heavy occlusion, Circle-2 is able to activate more details. The feature layer with the best activation regions (red dotted ellipses) corresponds to the detection results.

Fig.\ \ref{Fig.loss} shows the classification and regression loss of Circle-1 and Circle-2 on the predictors. We can see that the loss of Circle-2 is larger than that of Circle-1. The reason lies in that during the feature adaptation procedure, Circle-2 focuses on learning the features for harder instances and the features of Circle-2 inherit the Circle-1, as analyzed below.

\textbf{Instance Decomposition:}
We study the effect of different instance decomposition strategies on the circles. ``None" means that Circle-1 and Circle-2 are trained with all instances. ``By loss" means the instances are decomposed according to the loss, which requires Circle-2 to learn the hard instances with larger loss in the training batch. The RoIs of larger training loss in Circle-2 are defined as hard instances. We normalize the loss of instances in a training batch and scale them to a specific range. The final weights are obtained by adding them to a basic value as
$w=\frac{l-l{min}}{l{max}-l{min}} \times (1-\alpha) + \alpha $, where $l$ means the loss of the training batch, and we empirically set $\alpha=0.7$. 
This strategy improves the performance on high-resolution sub-set (Height $\geq$ 50) by 0.49\% (22.37\% vs. 22.86\%), while maintaining the performance on the sub-set (Height $\geq$20) (Table\ \ref{TabCaltechAblation2}). 

\begin{figure}[t]
    \centering
    \includegraphics[width=0.7\linewidth]{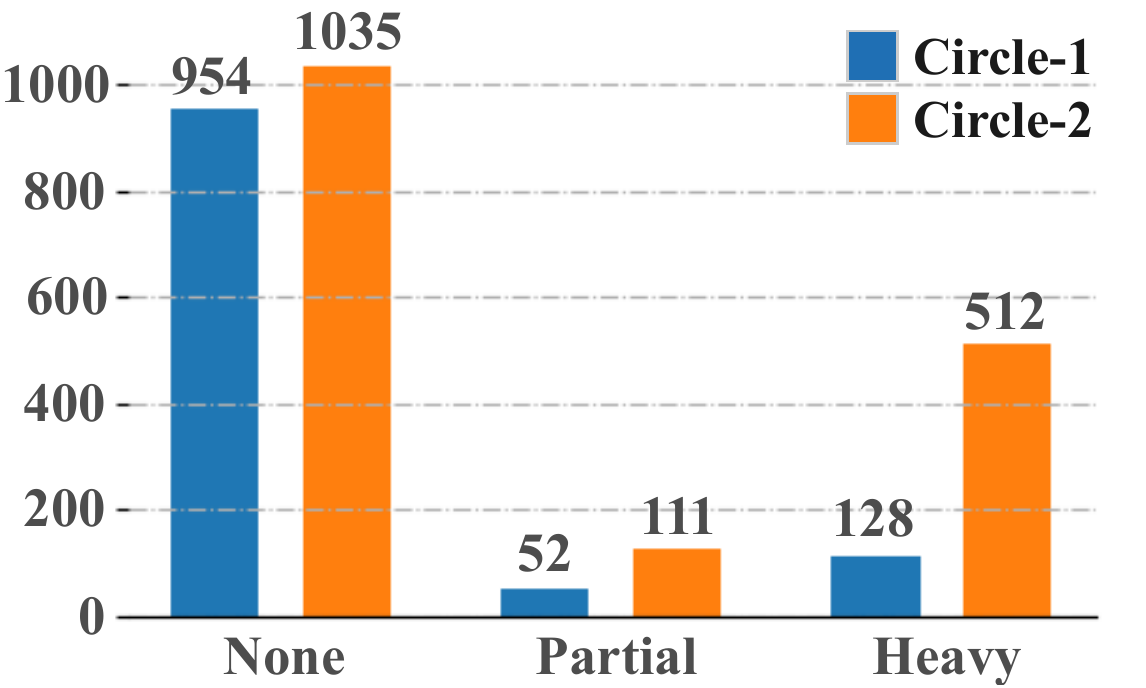}
    \caption{\label{Fig.Instance_decomposition_bar} Detection result statistics. Circle-1 and Circle-2 detect a comparable number of objects with occlusion ratio ``None" and ``Partial", but Circle-2 detects significantly more ``Heavy" occlusion objects. }
\end{figure}

Pedestrians with an occlusion ratio between $65\%$ and $80\%$ are defined as hard instances. ``All-to-hard" means all samples are used in Circle-1, and ``hard" samples are learned in Circle-2.  ``Easy-to-hard" means easy samples are decomposed in the first circle, and ``Easy+hard" samples are decomposed to the second one. As shown in Table\ \ref{TabCaltechAblation2}, ``Easy-to-hard" is the best instance decomposition strategy, which reduces the error rate $\mathrm{MR^{-2}}$ by $1.85\%$ (52.69\% vs. 54.54\%) on the ``All" (Height$\geq$20) sub-set.

\begin{figure}[tbp]
    \centering
    \subfloat[]{\label{Fig.t-SNE:subfig_a}
    \begin{minipage}{0.5\linewidth}
        \centering
        \includegraphics[width=0.7\textwidth]{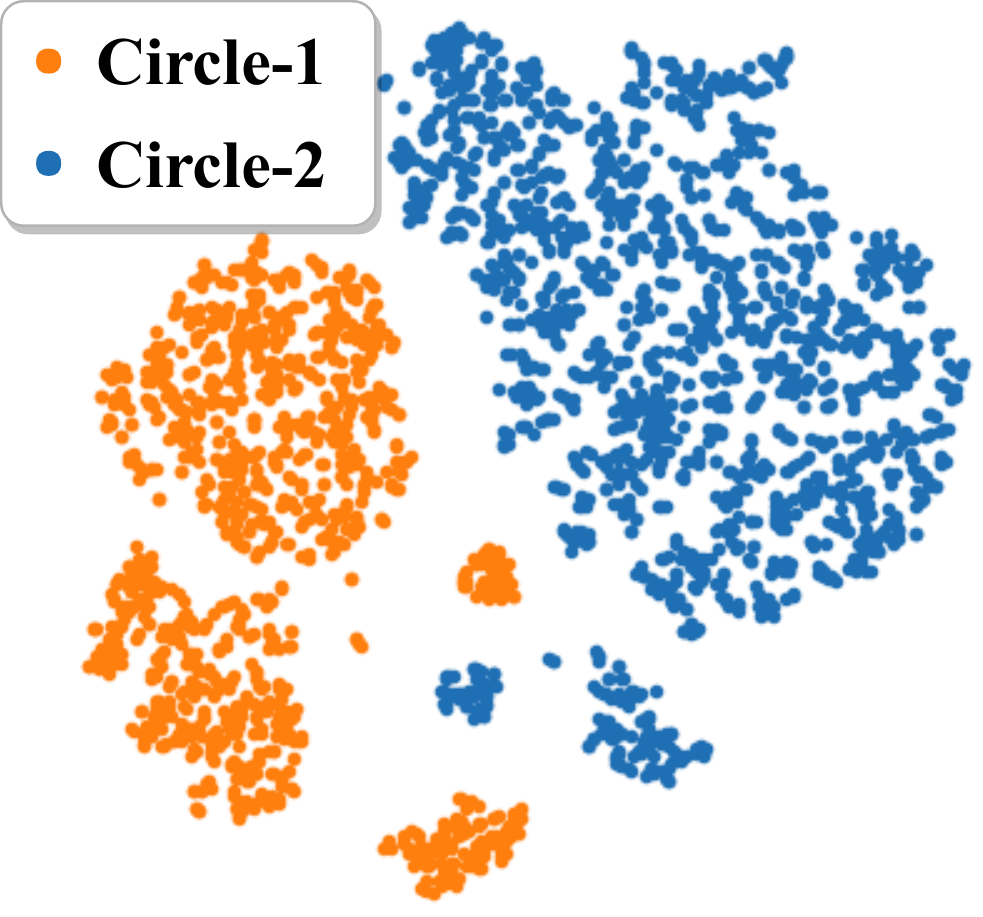}
    \end{minipage}}
    \subfloat[]{\label{Fig.t-SNE:subfig_b}
    \begin{minipage}{0.5\linewidth}
        \centering
        \includegraphics[width=0.8\textwidth]{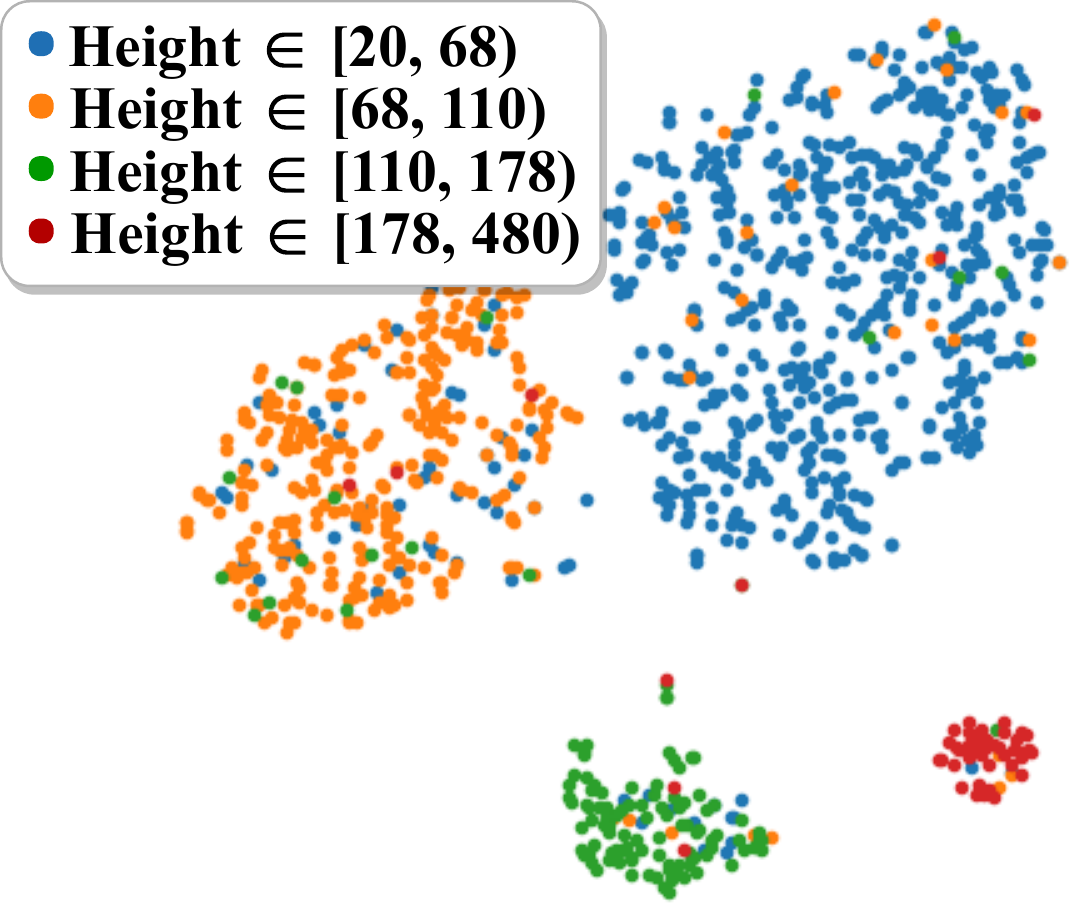}
    \end{minipage}}
    \caption{\label{Fig.t-SNE} The t-SNE \cite{DBLP:journals/ml/MaatenH12} \cite{DBLP:journals/jmlr/Maaten14} visualization of different feature embedding. (a) Instances from different circles. (b) Instances with different resolution. (Best view in color)}
\end{figure}

\begin{figure}[tbp]
    \subfloat[Positive examples]{\label{Fig:Mask:subfig_a}
    \begin{minipage}{0.5\linewidth}
        \includegraphics[width=4.1cm]{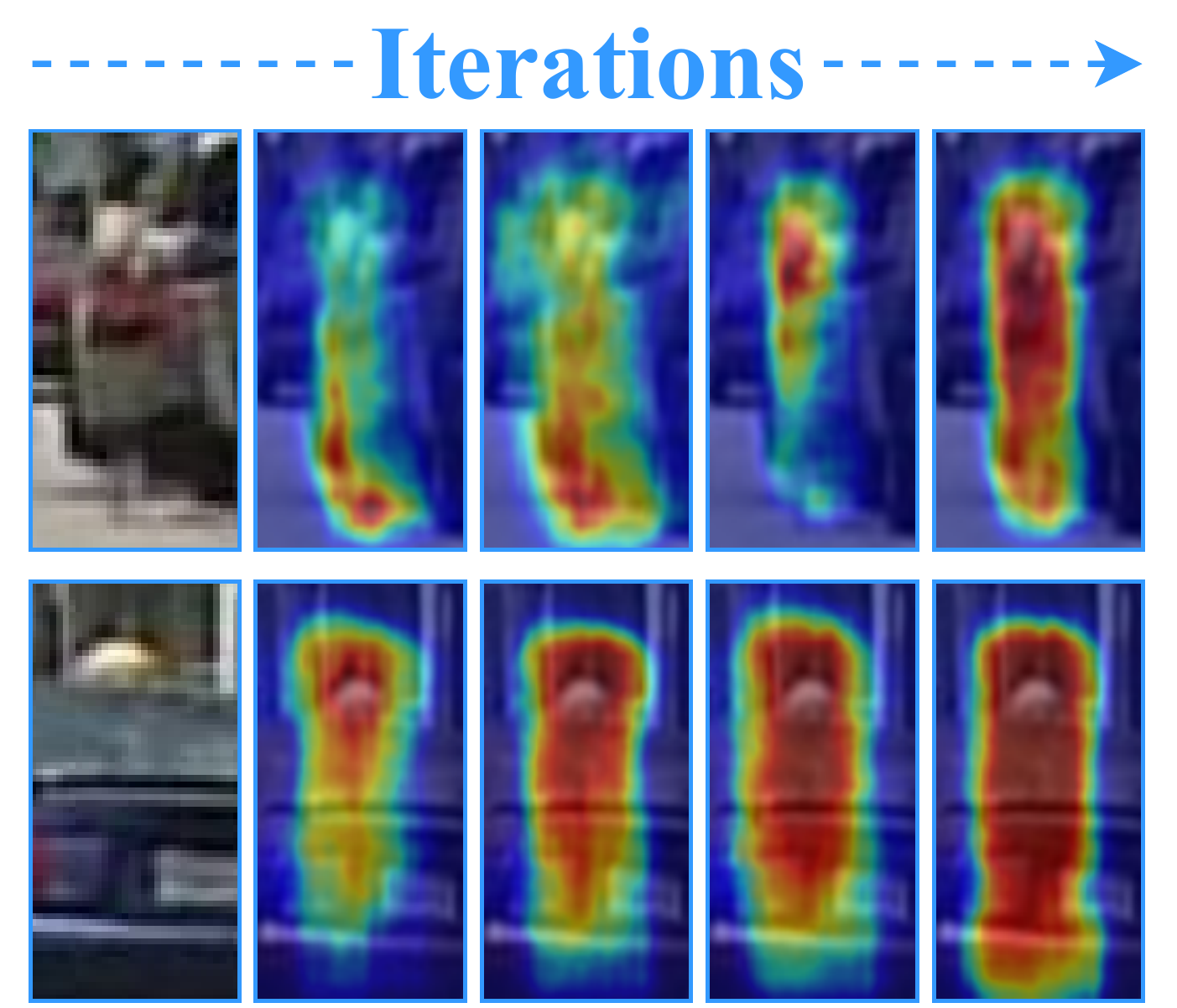}
    \end{minipage}}
    \subfloat[Negative examples]{\label{Fig.Mask:subfig_b}
    \begin{minipage}{0.5\linewidth}
        \includegraphics[width=4.1cm]{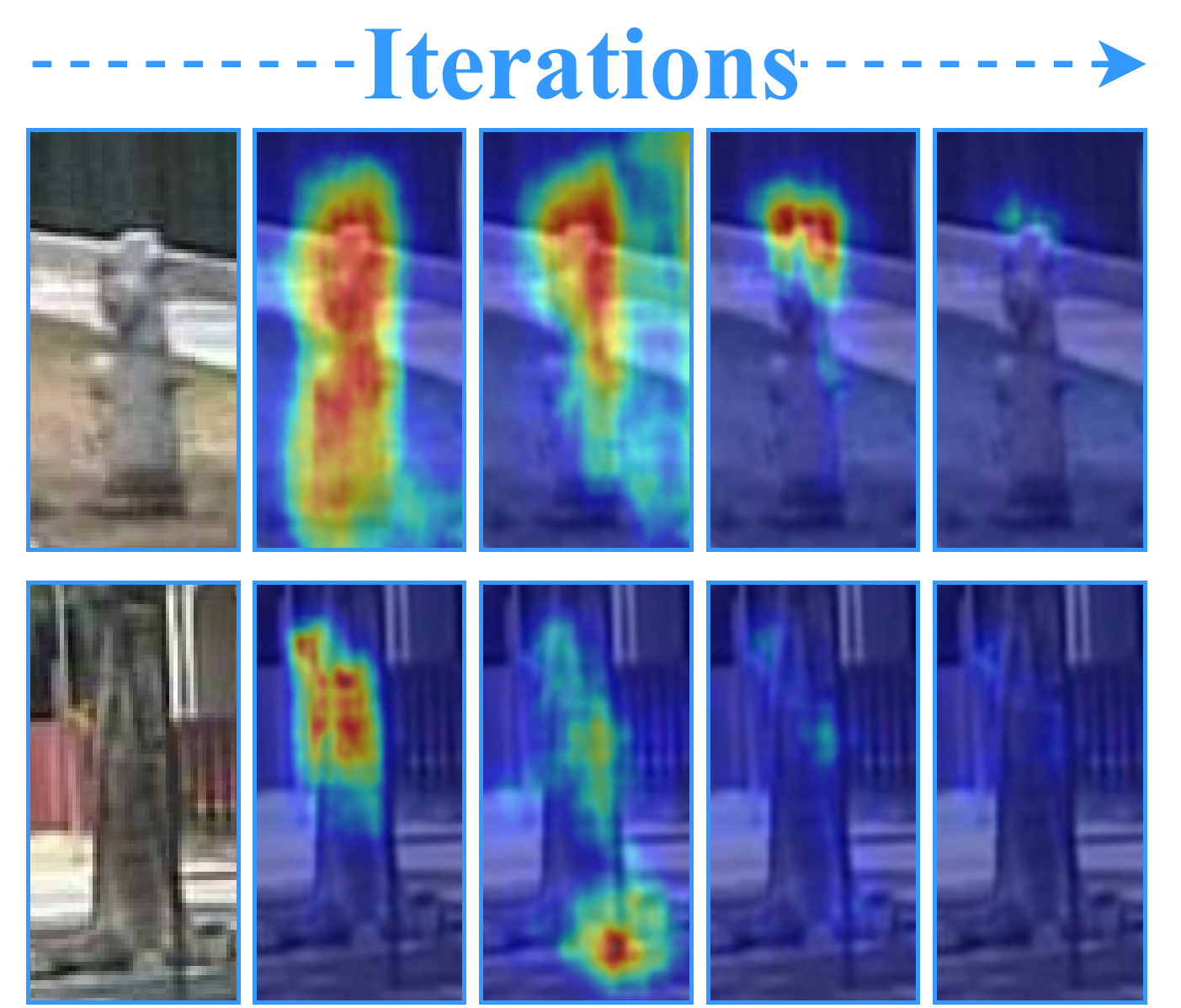}
    \end{minipage}}
    \caption{\label{Fig.Mask} With multiple supervisions, the feature representation for the positive examples are enforced while that for the negative examples are depressed. (Best viewed in color)}
\end{figure}

\begin{figure}[tbp]
    \subfloat[CircleNet]{\label{fig:subfig_a}
    \begin{minipage}{1\linewidth}
        \centering
        \includegraphics[width=8.4cm,height=2cm]{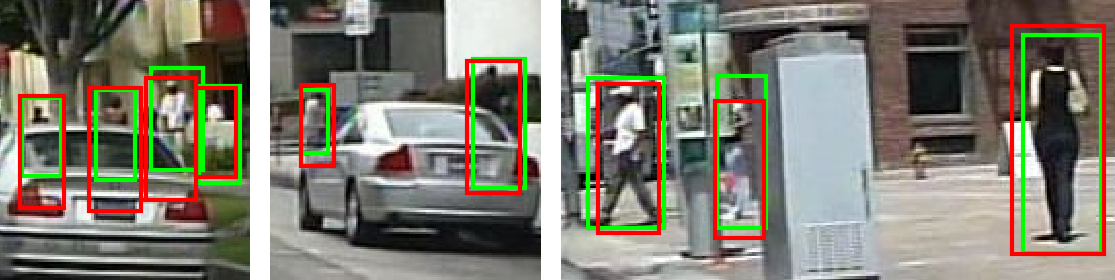}
    \end{minipage}}

    \subfloat[FasterRCNN+ATT \cite{zhang2018occluded}]{\label{fig:subfig_b}
    \begin{minipage}{1\linewidth}
        \centering
        \includegraphics[width=8.4cm,height=2cm]{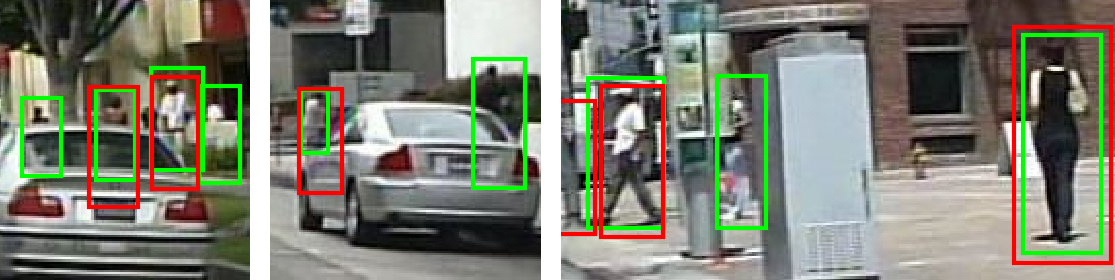}
    \end{minipage}}

    \subfloat[MS-CNN \cite{DBLP:conf/eccv/CaiFFV16}]{\label{fig:subfig_c}
    \begin{minipage}{1\linewidth}
        \centering
        \includegraphics[width=8.4cm,height=2cm]{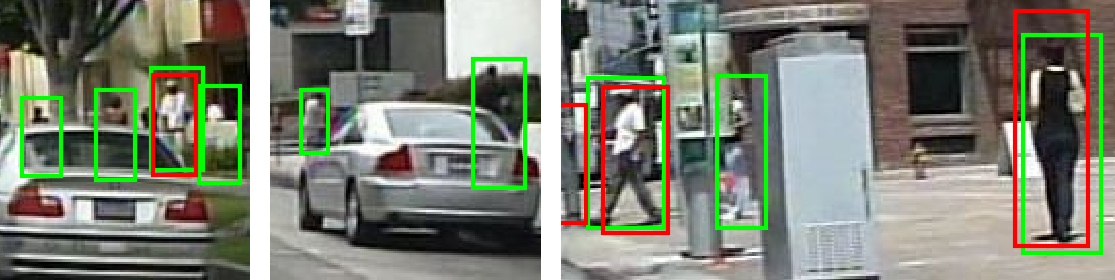}
    \end{minipage}}
    
    \caption{\label{Fig.detections} Qualitative detection results of cropped image patches at FPPI=0.1 on the Caltech test set. The green solid boxes indicate ground truth; the red boxes denote detection results. (Best viewed in color)}
    \label{Fig.4}
\end{figure}

In the test phase, we randomly select 2792 pedestrian samples from the test dataset, and plot the detection result statistics according to occlusion ratio (None, Partial and Heavy) in Fig.\ \ref{Fig.Instance_decomposition_bar}. It can be seen that both Circle-1 and Circle-2 have detected a comparable number of objects with occlusion ratio ``None" and ``Partial", but Circle-2 has detected significantly more ``Heavy" occlusion objects.

\begin{table*}[]
\centering
\setlength{\tabcolsep}{2.7mm}
\fontsize{7}{9}\selectfont
\begin{tabular}{c|c|ccccc|cccc}
\hline
\multirow{2}{*}{Model} & \multirow{2}{*}{Instance decomposition \& supervision} & \multicolumn{5}{c|}{Height$\geq${}50} & \multicolumn{4}{c}{Height$\geq${}20} \\ \cline{3-11}
                       &                           & All   & None  & Partial & Heavy & Reasonable & All       & None     & Partial    & Heavy    \\ \hline \hline
CircleNet              & w/o                                  & 24.14 & 12.84 & 28.74   & 54.66 & 15.02      & 55.36     & 46.97    & 65.77      & 75.05    \\ \hline
CircleNet+ID1       & None                & 22.86 & 12.33 & 27.68   & 50.4  & 14.54      & 54.54     & 47.42    & 66.11      & 69.42    \\
CircleNet+ID2       & By loss (OHEM)\cite{DBLP:conf/cvpr/ShrivastavaGG16}        & 22.37 & 12.92 & 28.45   & 48.35 & 14.83      & 54.55     & 47.15    & 65.31      & 72.25    \\
CircleNet+ID3       & All-to-hard            & 23.84 & 12.25 & 28.00   & 55.19 & 14.48      & 55.05     & 46.40    & 65.65      & 76.48    \\
CircleNet+ID4       & Easy-to-hard            & 21.62 & 11.83 & 26.45   & 48.70 & 13.78      & 52.69     & 44.29    & 64.83      & 73.09    \\
CircleNet+MS        & Multiple supervision                     & 21.57 & 12.44 & 27.50   & 45.47 & 14.38      & 54.62     & 47.12    & 66.29      & 71.55    \\ \hline
CircleNet+          & Multiple supervision + decomposition & 18.05 & 8.42  & 20.27   & 44.53 & 10.21      & 46.42     & 37.48    & 59.26      & 66.28    \\ \hline
\end{tabular}
\caption{Detection performance of CircleNet with instance decomposition and multiple supervisions on the Caltech test set.}
\label{TabCaltechAblation2}
\end{table*}

\begin{table*}[]
\centering
\setlength{\tabcolsep}{4.4mm}
\fontsize{7}{9}\selectfont
\begin{tabular}{c|ccccc|cccc}
\hline
\multirow{2}{*}{Model} & \multicolumn{5}{c|}{Height$\geq${}50}                                                                   & \multicolumn{4}{c}{Height$\geq${}20}                                              \\ \cline{2-10}
                       & All                  & None                & Partial              & Heavy                & Reasonable          & All                  & None                 & Partial              & Heavy                \\ \hline \hline
DeepParts \cite{DBLP:conf/iccv/TianLWT15} & 22.79                & 10.64               & 19.93                & 60.42                & 11.89               & 64.78                & 58.43                & 70.39                & 81.81                \\
MS-CNN \cite{DBLP:conf/eccv/CaiFFV16}  & 21.53                & 8.15                & 19.24                & 59.94                & 9.95                & 60.95                & 53.67                & 67.16                & 79.51                \\
RPN+BF \cite{DBLP:conf/eccv/ZhangLLH16}  & 24.01                & 7.68                & 24.23                & 74.36                & 9.58                & 64.66                & 56.38                & 72.55                & 87.48                \\
AdaptFasterRCNN \cite{DBLP:conf/cvpr/ZhangBS17} & 20.03                & 7.01                & 26.55                & 57.58                & 9.18                & 60.11                & 52.67                & 68.50                & 79.58                \\
SDS-RCNN \cite{DBLP:conf/iccv/BrazilYL17} & 19.72                & {\ul \textbf{5.95}}                & {\ul \textbf{14.86}}                & 58.55                & {\ul \textbf{7.36}} & 61.50                & 54.45                & 66.46                & 78.78                \\
FasterRCNN+ATT \cite{zhang2018occluded} & 18.21                & 8.46                & 22.29                & 45.18                & 10.33               & 54.51                & 47.54                & 64.47                & 71.02                \\
CircleNet (Ours)           & {\ul \textbf{18.05}} & 8.42 & 20.27 & {\ul \textbf{44.53}} & 10.21               & {\ul \textbf{46.42}} & {\ul \textbf{37.48}} & {\ul \textbf{59.26}} & {\ul \textbf{66.28}} \\ \hline
\end{tabular}

\caption{Comparison of CircleNet with other state-of-the-art methods on the Caltech test set.}
\label{TabCaltechComparison}
\end{table*}

\begin{table}[]
\centering
\setlength{\tabcolsep}{2.3mm}
\fontsize{7}{9}\selectfont
\begin{tabular}{c|c|ccc}
\hline
Method              & \textit{Reasonable} & \textit{Heavy*} & \textit{Partial*} & \textit{Bare*} \\ \hline \hline
Adapted Faster RCNN \cite{DBLP:conf/cvpr/ZhangBS17}   & 12.8                & -              & -                & -             \\
Repulsion Loss      \cite{wang2017repulsion}          & 11.6                & 55.3           & 14.8             & 7.0           \\
OR-CNN              \cite{DBLP:conf/eccv/ZhangWBLL18} & 11.0              & 51.3           & 13.7             & 5.9           \\
CircleNet (Ours)      & 11.77    & {\ul \textbf{50.22}}          & {\ul \textbf{12.21}}            & 7.14          \\ \hline
\end{tabular}
\caption{Comparison of CircleNet with other state-of-the-art methods on the CityPersons validation set. (*denotes the experimental protocols used in \cite{wang2017repulsion} \cite{DBLP:conf/eccv/ZhangWBLL18}).}
\label{TabCityPersonsComparison}
\end{table}

\begin{figure*}[tbp]
    \subfloat[All (Height$\geq${}50)]{\label{Fig.CaltechCurves:subfig_a}
    \begin{minipage}{0.30\linewidth} 
        \centering
        \includegraphics[width=6cm]{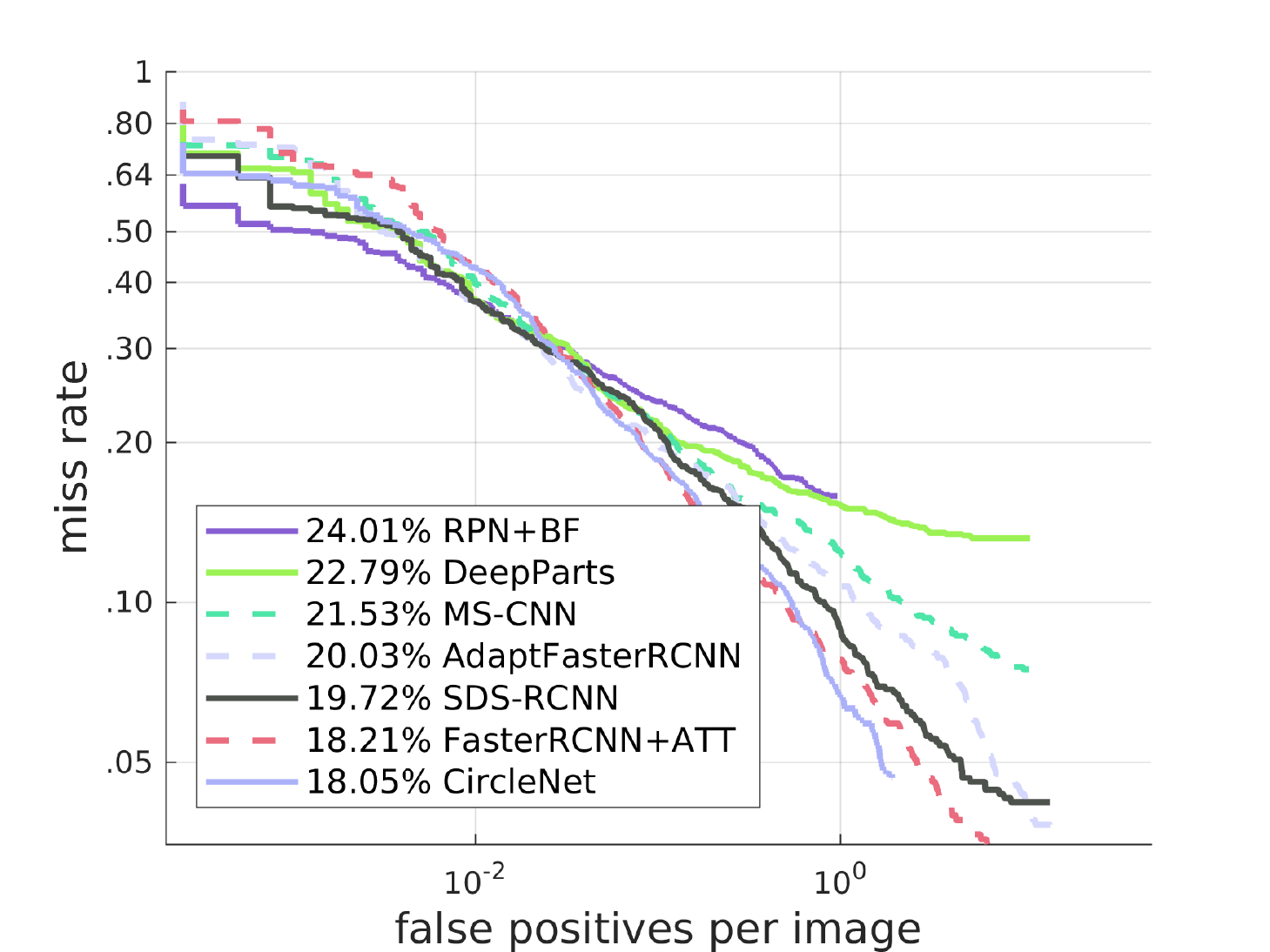}
    \end{minipage}}
    \subfloat[Heavy (Height$\geq${}50)]{\label{Fig.CaltechCurves:subfig_c}
    \begin{minipage}{0.30\linewidth}
        \centering
        \includegraphics[width=6cm]{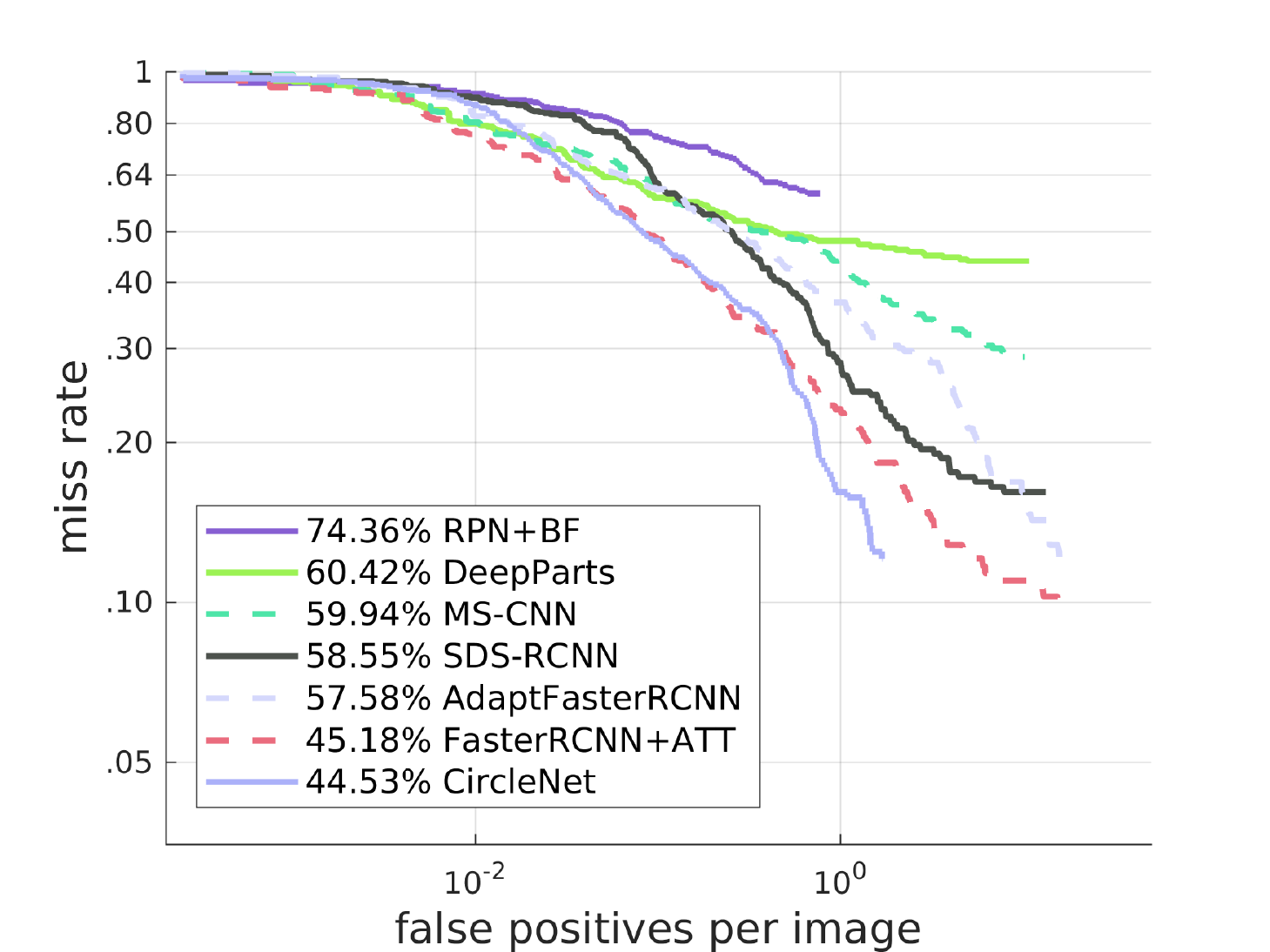}
    \end{minipage}}
    \subfloat[All (Height$\geq${}20)]{\label{Fig.CaltechCurves:subfig_b}
    \begin{minipage}{0.30\linewidth} 
        \centering
        \includegraphics[width=6cm]{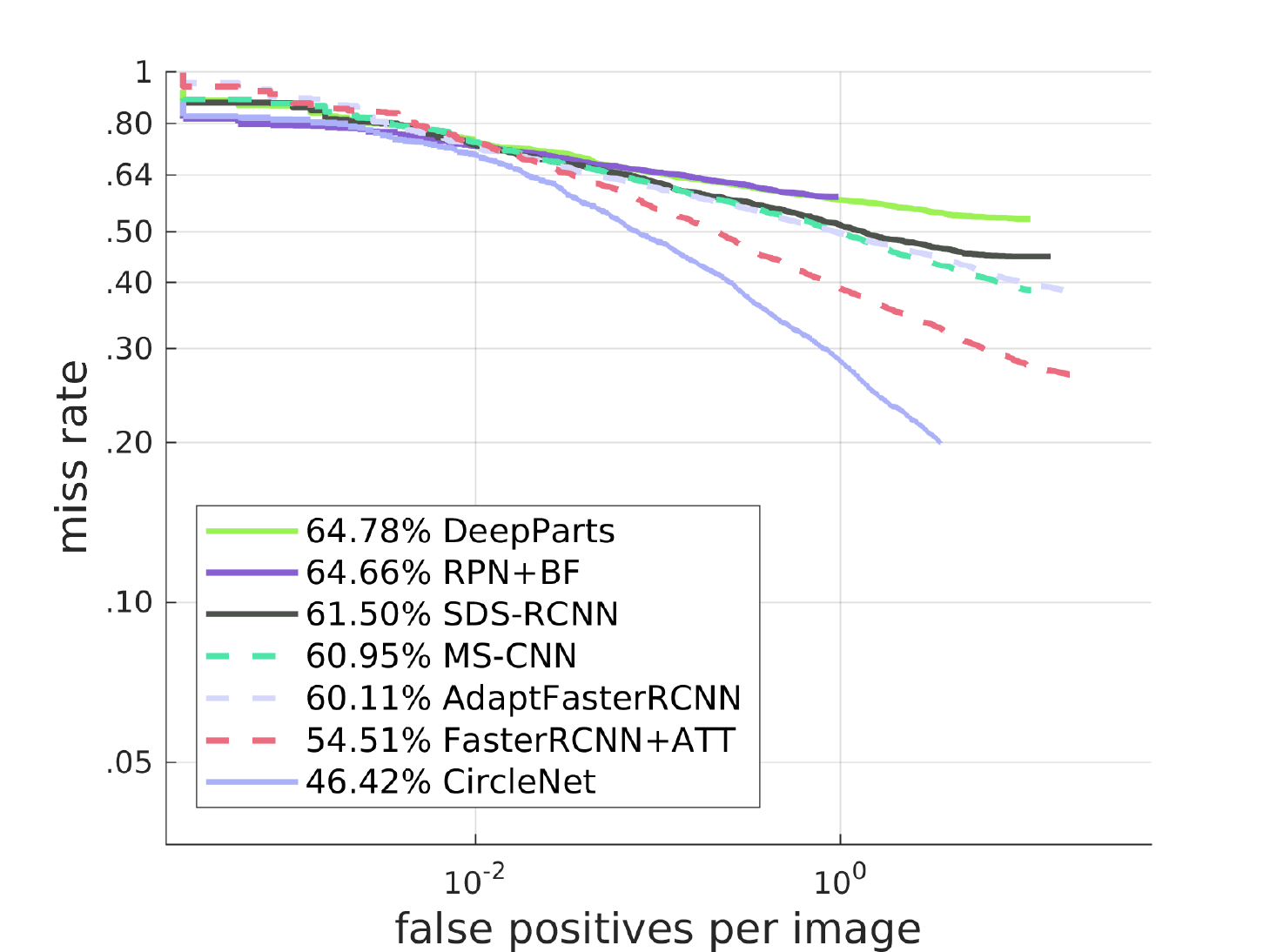}
    \end{minipage}}
    
    \subfloat[None (Height$\geq${}20)]{\label{Fig.CaltechCurves:subfig_d}
    \begin{minipage}{0.30\linewidth}
        \centering
        \includegraphics[width=6cm]{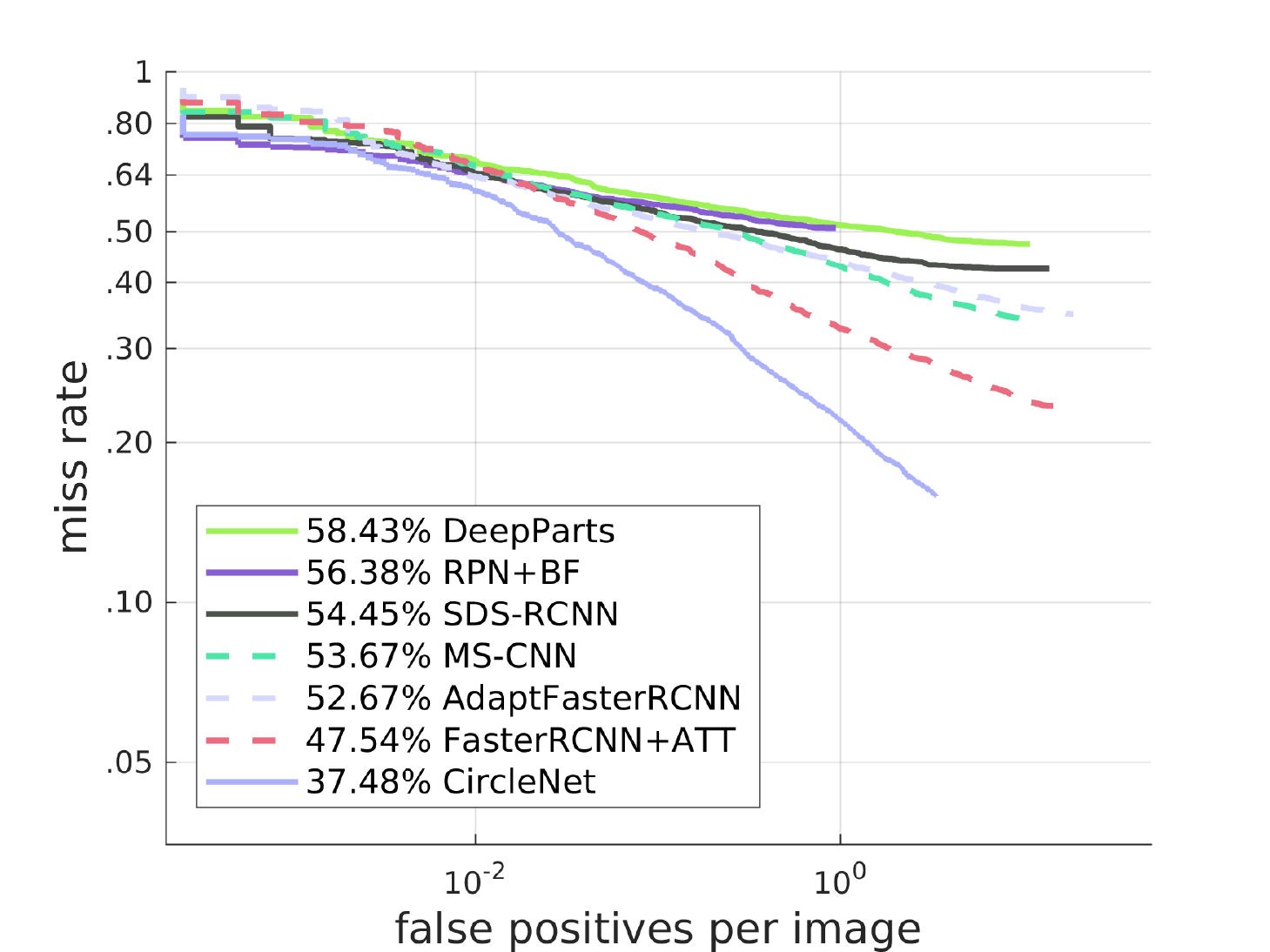}
    \end{minipage}}
    \subfloat[Partial (Height$\geq${}20)]{\label{Fig.CaltechCurves:subfig_d}
    \begin{minipage}{0.30\linewidth}
        \centering
        \includegraphics[width=6cm]{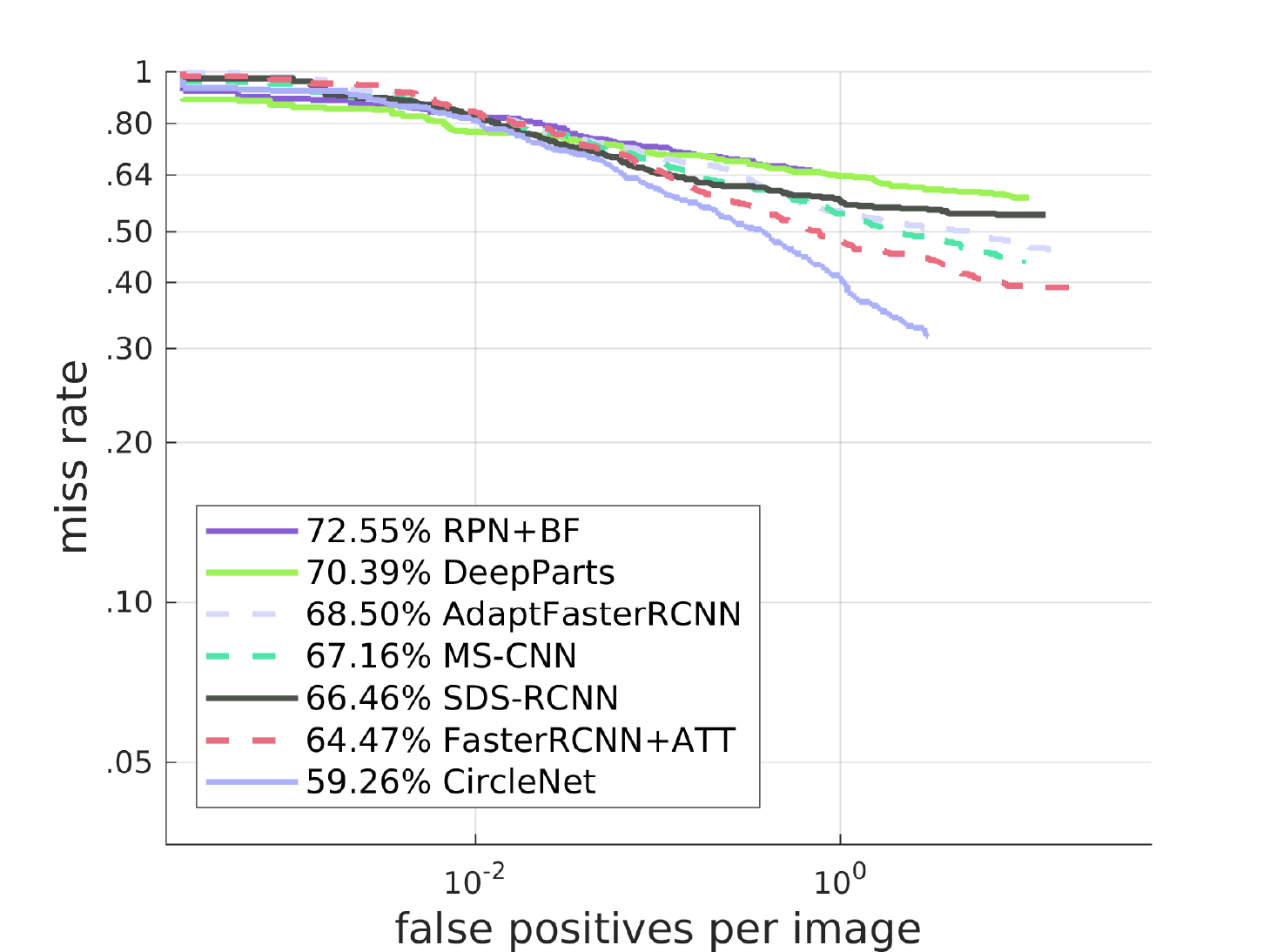}
    \end{minipage}}
    \subfloat[Heavy (Height$\geq${}20)]{\label{Fig.CaltechCurves:subfig_d}
    \begin{minipage}{0.30\linewidth}
        \centering
        \includegraphics[width=6cm]{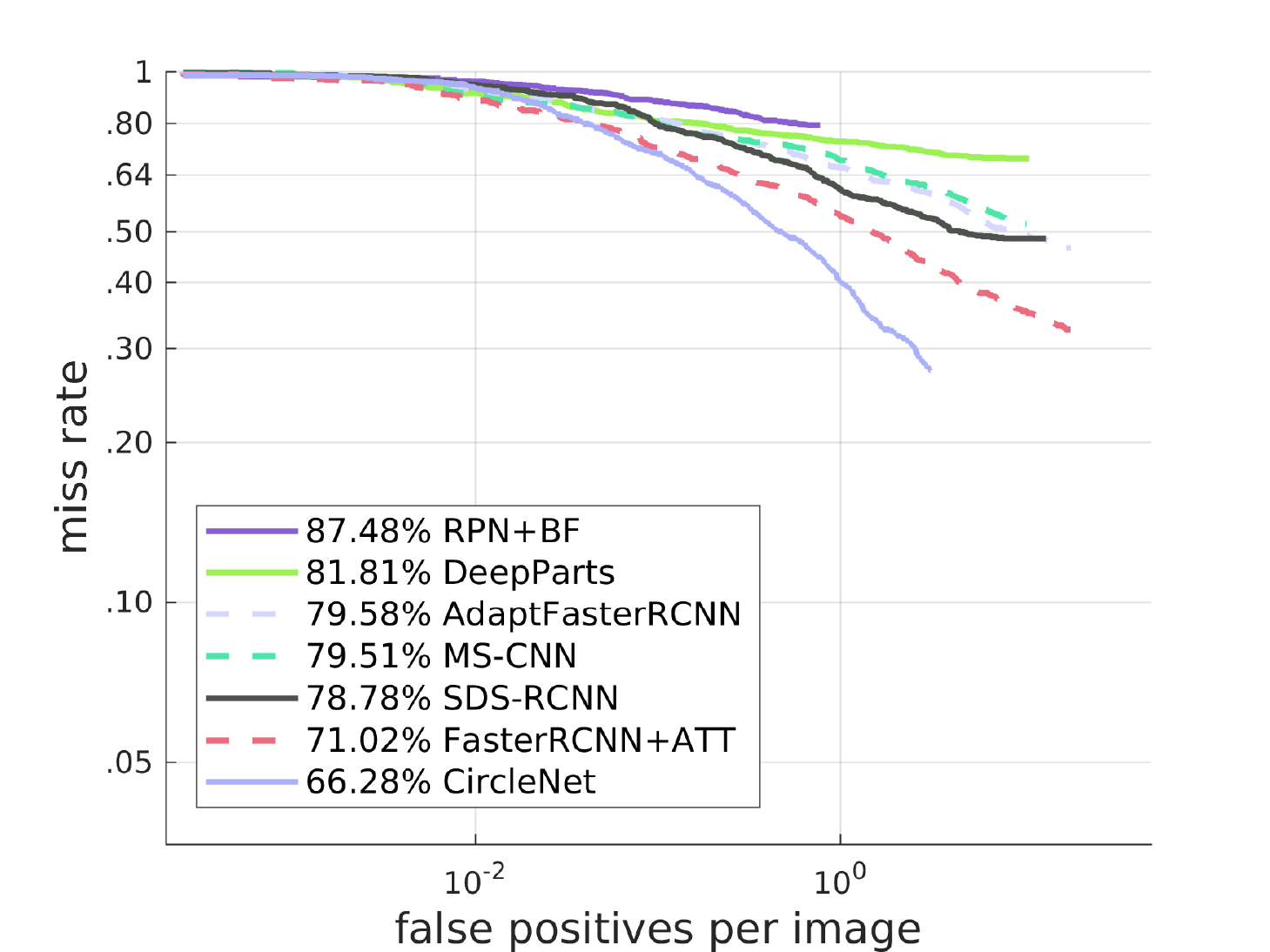}
    \end{minipage}}
    \caption{Performance comparison on the Caltech dataset. Lower curves indicate better performance.} 
    \label{Fig.CaltechCurves}
\end{figure*}

\begin{figure}[t]
    \centering
    \includegraphics[width=1.0\linewidth]{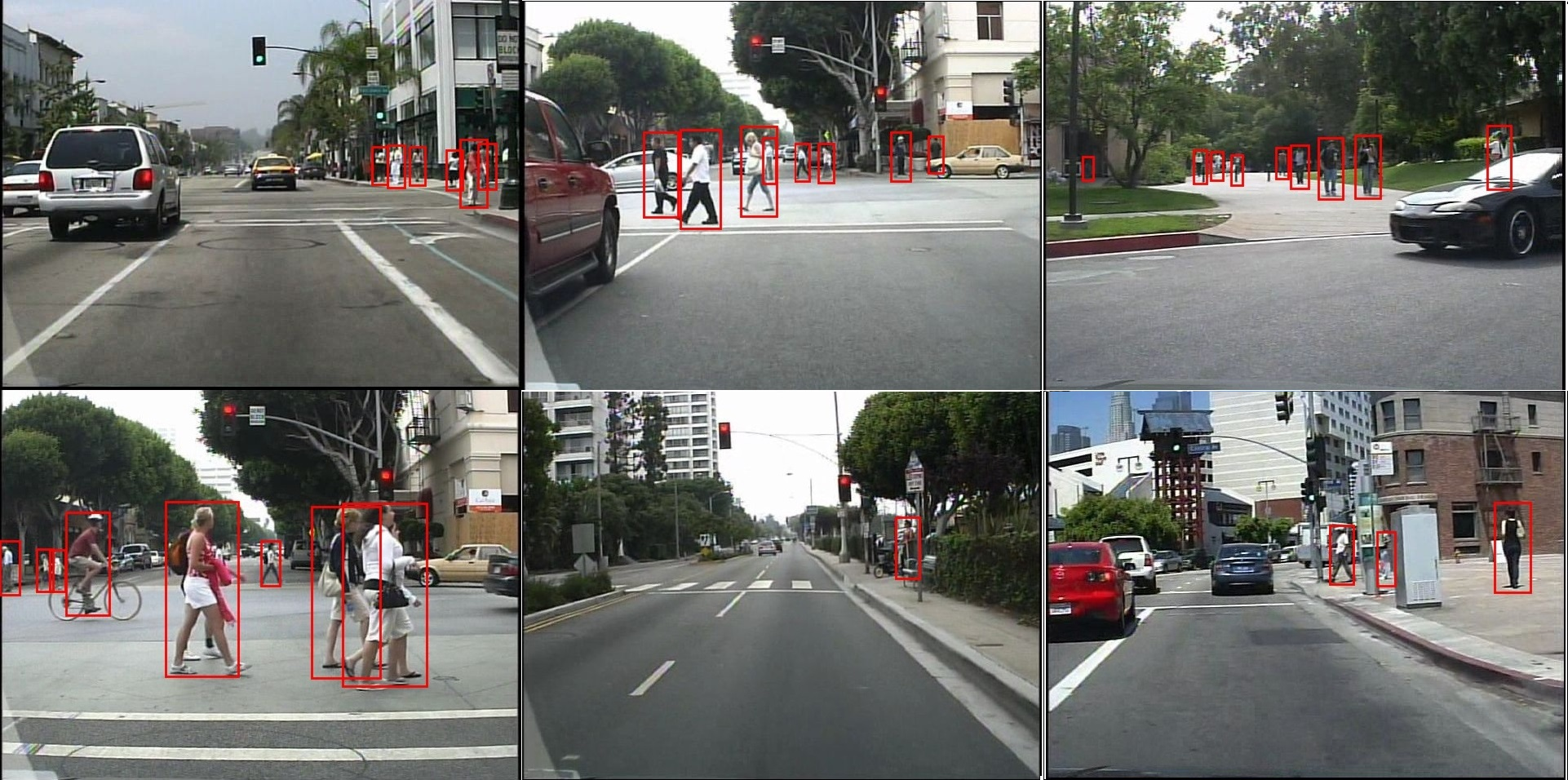}
    \caption{\label{Fig.Caltech_Detections} Detection examples from the Caltech dataset. Red bounding boxes are predicted pedestrians with threshold 0.7.}
\end{figure}
\begin{figure}[t]
    \centering
    \includegraphics[width=1.0\linewidth]{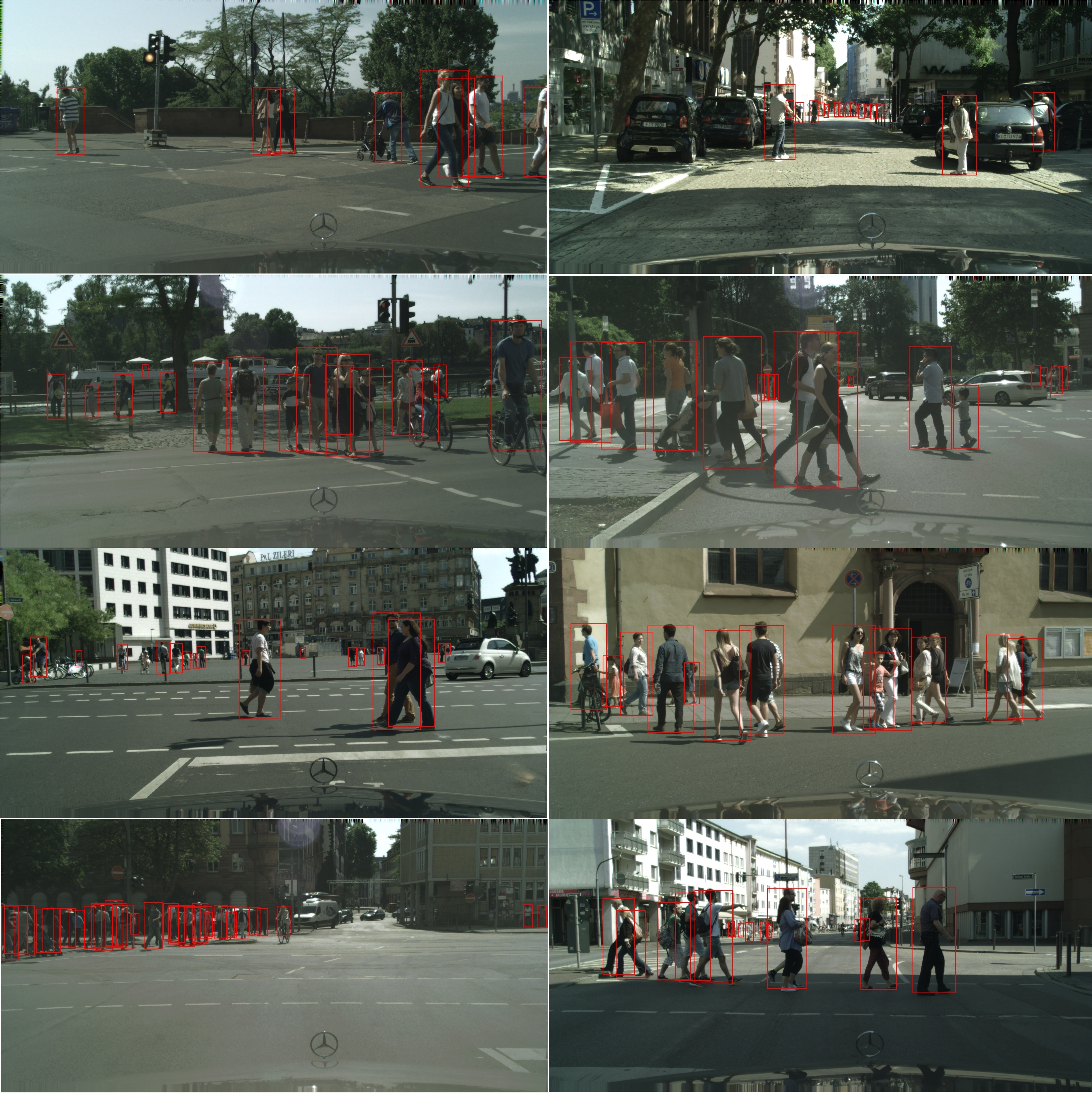}
    \caption{\label{Fig.CityPersons_Detections} Detection examples from the CityPersons dataset. Red bounding boxes are predicted pedestrians with threshold 0.7. (Best viewed in color and with zoom.)}
\end{figure}

We further visualize the feature embedding of these pedestrian samples using t-SNE \cite{DBLP:journals/ml/MaatenH12} \cite{DBLP:journals/jmlr/Maaten14}, Fig.\ \ref{Fig.t-SNE}. It can be seen that the samples from Circle-1 and Circle-2 form clusters, which correspond to pedestrian instances of different appearances in Fig.\ \ref{Fig.t-SNE:subfig_a}. This figure shows the feature differences between adapted circles. Fig.\ \ref{Fig.t-SNE:subfig_b} shows the clusters of samples in terms of resolution. CircleNet, as a classifier ensemble, can process these clusters well.

\textbf{Multiple Supervision:}
With object detection loss and pseudo-segmentation loss, we target at activating features on the object border while further explore the feature adaptability of CircleNet. Fig.\ \ref{Fig:Mask:subfig_a} 
illustrates that when training, the activation mask \footnote{The activation mask is obtained by a segmentation layer ($3\times3$ and $1\times1$ convolutional operations) and a sigmoid function.} gradually fills the bounding boxes. As a pixel-wise classification task, semantic segmentation with pseudo-mask supervision helps suppressing the negative samples from cluttered backgrounds, Fig.\ \ref{Fig.Mask:subfig_b}. 

\subsection{Performance and Comparison}

In Table\ \ref{TabCaltechComparison}, we compare the performance of  CircleNet with state-of-the-art approaches on Caltech. MS-CNN \cite{DBLP:conf/eccv/CaiFFV16}, RPN+BF \cite{DBLP:conf/eccv/ZhangLLH16}, AdaptFaster-RCNN \cite{DBLP:conf/cvpr/ZhangBS17}, and SDS-RCNN \cite{DBLP:conf/iccv/BrazilYL17} achieve top results on the ``Reasonable" sub-set, but do not perform well on heavily occluded instance or low-resolution instances. CircleNet beats all compared approaches on low-resolution and occluded cases, while reporting acceptable performance on ``Reasonable" sub-set.

As shown in Table~\ref{TabCaltechComparison}, CircleNet outperforms the state-of-the-art methods on the Caltech test set. It outperforms FastRCNN-ATT by 2.02\% (20.27\% vs. 22.29\%) on the ``Partial" sub-set (Height $\geq$ 50) and 8\% (46.42\% vs. 54.51\%) on the ``all" sub-set (Height $\geq$ 20). CircleNet also outperforms FastRCNN-ATT on the “Partial” and “Heavy” occlusion sub-set with significant margins.

Fig.\ \ref{Fig.detections} shows qualitative results which indicates that CircleNet produces robust detections, which  are well aligned with the ground-truth on various occlusion patterns. In contrast, the other two detectors (FasterRCNN+ATT \cite{zhang2018occluded} and MS-CNN \cite{DBLP:conf/eccv/CaiFFV16}) missed some of the objects. In Fig.\ \ref{Fig.CaltechCurves}, the miss rate and FPPI curves show that the proposed CircleNet outperforms state-of-the-art approaches with significant margins.

In Table\ \ref{TabCityPersonsComparison}, we compare CircleNet with state-of-the-art detectors on the CityPersons validation set, using the performance reported by the authors \cite{wang2017repulsion} \cite{DBLP:conf/eccv/ZhangWBLL18}. It can be seen that CircleNet consistently outperforms the state-of-the-art OR-CNN and FasterRCNN+ATT on this validation set.  In Fig.\ \ref{Fig.Caltech_Detections} and Fig.\ \ref{Fig.CityPersons_Detections}, the results show that CircleNet can effectively detect pedestrians of low-resolution, occlusion, and clutter backgrounds.

\section{Conclusion}
Pedestrian detection in the wild remains a challenging problem for the ``hard" instances with heavy occlusion and/or low resolution. In this paper, we developed a new feature learning model, referred to as CircleNet, which reciprocates the feature adaptation by formatting deep-to-shallow and shallow-to-deep feature fusion pathways. These cycling loops not only improve the representative capability and adaptability of convolutional features for objects of various appearance, but also naturally mimic the process which we humans attempt to detect and recognize small and highly occluded objects. We also propose hard instance decomposition strategies to assign instances along feature layers and circles to fully utilize the feature adaptation capability introduced by the circling architecture. Significant performance improvement over the baseline FPN approach demonstrates that CircleNet is simple yet effective. Its shows great potential for single-camera based pedestrian detection systems and provides fresh insight to occluded and low-resolution object detection problems.

\section*{Acknowledgment}
The authors would like to express their sincere appreciation to the reviewers for their constructive comments. This work was supported in part by NSFC under Grant 61836012, 61671427 and 61771447, and Beijing Municipal Science and Technology Commission under Grant
Z181100008918014.

\ifCLASSOPTIONcaptionsoff
  \newpage
\fi

\bibliographystyle{IEEEtran}
\bibliography{egbib}
%

\end{document}